\documentclass{article}

\PassOptionsToPackage{numbers, compress}{natbib}

\usepackage[final]{neurips_data_2024}

\usepackage[dvipsnames]{xcolor}
\usepackage[utf8]{inputenc}
\usepackage[T1]{fontenc}
\usepackage[colorlinks = true,
            linkcolor = BlueViolet,
            urlcolor  = BlueViolet,
            citecolor = BlueViolet,
            anchorcolor = BlueViolet]{hyperref}
\usepackage{url}
\usepackage{booktabs}
\usepackage{amsfonts}
\usepackage{enumitem}
\usepackage{nicefrac}
\usepackage{microtype}
\usepackage{xcolor}
\usepackage{graphicx}
\usepackage{bm}
\usepackage{amsmath}
\usepackage{cleveref}
\usepackage{multirow}
\usepackage{xspace}
\usepackage{pdflscape}
\usepackage{wrapfig}
\usepackage{ulem}
\usepackage{booktabs}

\setlist[itemize]{noitemsep, topsep=5pt}

\newcommand{\algo}{\texttt{CleanDiffuser}}

\newcommand{\alg}{\algo\xspace}
\newcommand{\dd}{{\rm d}}
\newcommand{\scorefunc}{$\nabla_{\bm x}\log q_t(\bm x_t)$}

\title{\alg: An Easy-to-use 
 Modularized Library for Diffusion Models in Decision Making}

\author{Zibin Dong$^1$\thanks{These authors contribute equally to this work.},~~Yifu Yuan$^1$\footnotemark[1],~~Jianye Hao$^1$\thanks{Corresponding authors:  Jianye Hao (jianye.hao@tju.edu.cn)},~~Fei Ni$^1$,~~Yi Ma$^2$,~~Pengyi Li$^1$,~~Yan Zheng$^1$ \\
$^1$College of Intelligence and Computing, Tianjin University\\
\texttt{\{zibindong,yuanyf,jianye.hao,fei\_ni,lipengyi,yanzheng@tju.edu.cn\}}\\
$^2$School of Computer and Information Technology, Shanxi University, \texttt{mayi@sxu.edu.cn}
}

\begin{document}

\maketitle
\begin{abstract}\label{sec:abstract}
Leveraging the powerful generative capability of diffusion models (DMs) to build decision-making agents has achieved extensive success. However, there is still a demand for an easy-to-use and modularized open-source library that offers customized and efficient development for DM-based decision-making algorithms. In this work, we introduce \textbf{\alg}, the first DM library specifically designed for decision-making algorithms. By revisiting the roles of DMs in the decision-making domain, we identify a set of essential sub-modules that constitute the core of \alg, allowing for the implementation of various DM algorithms with simple and flexible building blocks. To demonstrate the reliability and flexibility of \alg, we conduct comprehensive evaluations of various DM algorithms implemented with \alg across an extensive range of tasks. The analytical experiments provide a wealth of valuable design choices and insights, reveal opportunities and challenges, and lay a solid groundwork for future research. \alg will provide long-term support to the decision-making community, enhancing reproducibility and fostering the development of more robust solutions. 
The code and documentation of \alg are open-sourced on the \href{https://github.com/CleanDiffuserTeam/CleanDiffuser}{project website}.
\end{abstract}

\section{Introduction}\label{sec:introduction}

Diffusion models~(DMs)~\cite{ho2020ddpm, kingma2021variational, song2021scorebased} have emerged as a leading class of generative models, outperforming previous methods~\cite{creswell2018generative, kingma2013auto} in both high-quality generation and training stability~\cite{zhu2023diffusion}. Their remarkable capabilities in complex distribution modeling and conditional generation demonstrate promising performance across various domains~\cite{zhang2023controlnet, Ruiz2023dreambooth, Liu2023audioldm, kingma2021variational}, inspiring a series of works to apply DMs in decision-making tasks~\cite{xian2023chaineddiffuser, yang2023learning, dong2023aligndiff, wang2023diffusion, hansen2023idql, chi2023diffusion}. Open-source libraries can quantify progress in this emerging field, enable researchers to better understand and compare algorithm details, and promote the application of DMs. Currently, several high-quality libraries are available for DMs, such as Diffusers~\cite{von-platen-etal-2022-diffusers} and Stable Diffusion~\cite{rombach2021highresolution}, which provide exemplary designs for the computer vision and multimedia. However, support for decision-making is lacking. Although some pioneering research~\cite{chi2023diffusion, janner2022planning, ajay2022conditional} on DMs for decision-making has provided excellent codes, their algorithm-specific mechanisms and tightly coupled system architectures are not conducive to customized development. 

\begin{figure}[h]
\centering
\includegraphics[width=0.99\textwidth]{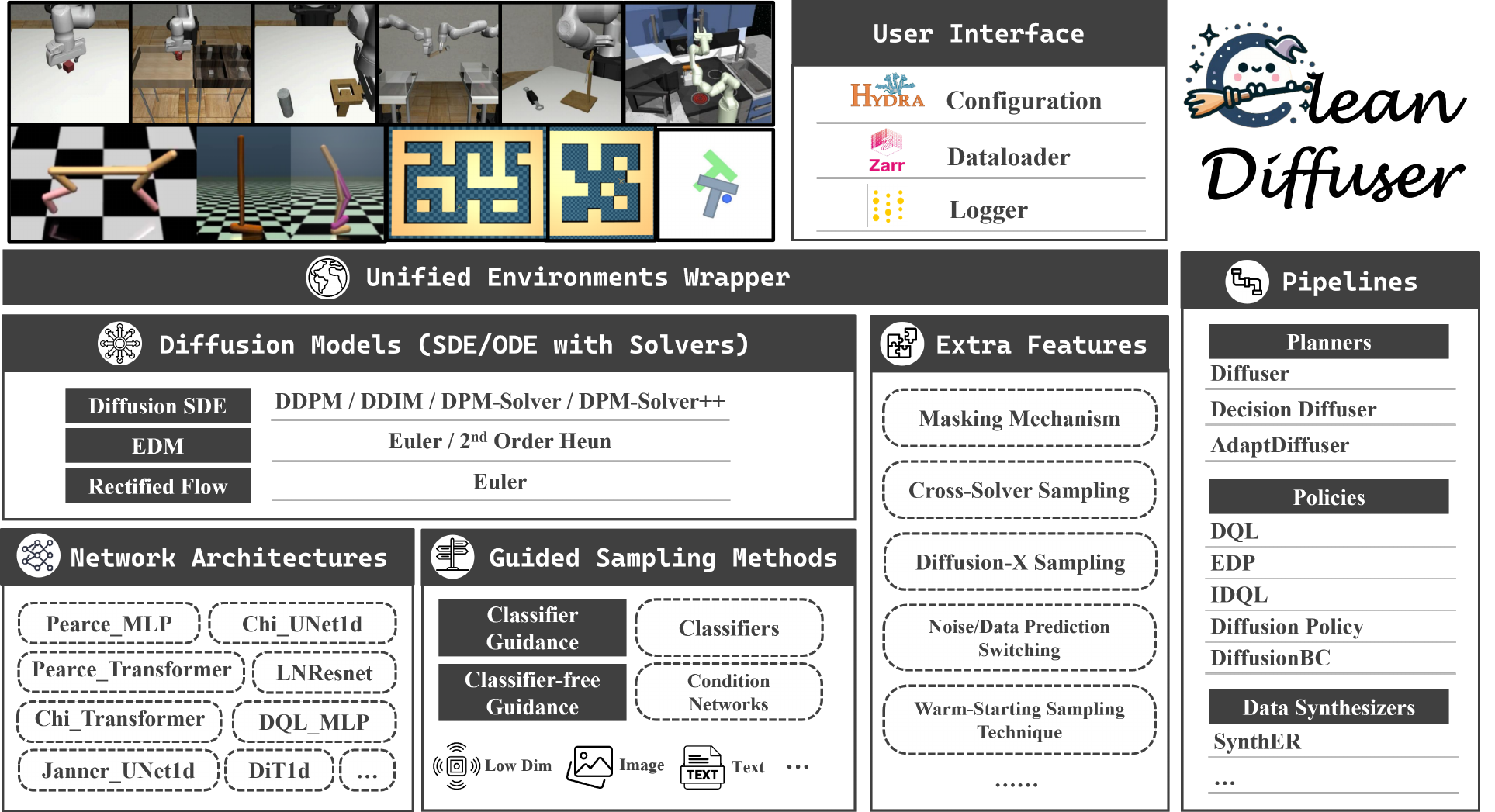}
\vspace{-5pt}
\caption{\small{\textbf{The Architecture of \alg.} \alg is specifically tailored for the decision-making domain, supporting a wide range of \texttt{Diffusion Models, Network Architectures}, and \texttt{Guided Sampling Methods} modules and extra useful features. By simply combining the building blocks into a \texttt{pipeline}, \alg integrates 9 popular DM algorithms.}}
\label{fig:framework}
\vspace{-10pt}
\end{figure}

In this paper, we present an easy-to-use modularized DM library tailored for decision-making named \alg, which comprehensively integrates different types of DM algorithmic branches. 
We revisit various roles of DMs in decision-making tasks and identify core sub-modules: \texttt{Diffusion Models, Network Architectures} and \texttt{Guided Sampling Methods}. \alg also incorporates an efficient \texttt{Dataloader} and useful \texttt{Environment Wrappers} for easy usage and customized datasets extension. Specifically, to address the unique decision-making challenges, \alg designs a series of practical features for special mechanisms. With \alg, algorithms can be implemented by selecting building blocks and integrating them into a pipeline. Customizing an algorithm requires only about 10 lines of code, providing the highest usability and customization. 
The decoupled modular architecture allows developers to adapt to different tasks and facilitates the adjustment of existing methods without complex abstractions. \alg effectively meets the diverse requirements of various decision-making algorithms.

To demonstrate the reliability and flexibility of \alg, we conduct extensive experiments in 37 Reinforcement Learning~(RL) and Imitation Learning~(IL) environments for 9 algorithms and their variants, benchmarking performance for many DM algorithms and serving as valuable references for future research. Thanks to the general architecture of \alg, we revisit the key design choices of the DMs for decision-making from a unified perspective. We conduct extensive empirical analyses on different architectures, solvers, sample steps, EMA, and model sizes, providing valuable insights and showing challenges for designing DM-based decision-making algorithms. 

Our contributions are three-fold: (1) We present an easy-to-use modularized library named \alg, the first DM library designed specifically for decision-making tasks. (2) We decouple the general DM algorithms into 3 core sub-modules and design specialized features for decision-making, ultimately integrating them into a modular pipeline. (3) Utilizing over 30,000 GPU hours of computational resources, we benchmark various popular DM-based algorithms and conduct a thorough empirical analysis, providing valuable insights and revealing opportunities and challenges. 

\section{Background}

\textbf{Sequential Decision-making Problem.} Consider a system governed by discrete-time dynamics $(\bm s^{t+1}, r^t)=d(\bm s^t,\bm a^t)$, in which taking action $\bm a^t$ at state $\bm s^t$ leads a transition to $\bm s^{t+1}$ and yields a scalar reward $r^t$. Given an interaction record dataset $\mathcal D=\{(\bm s^t,\bm a^t,r^t,\bm s^{t+1})\}$ collected by a behavior policy, the offline RL~\citep{fu2020d4rl, fujimoto2018addressing} aims to derive an optimal policy from the dataset to maximize cumulative reward and surpass the behavior policy. The offline IL~\citep{mandlekar2020learning}, which assumes the behavior policy is an expert and does not require reward labels, aims to mimic the expert behaviors closely.

\textbf{Training and Sampling of Diffusion Models.} Assume a $D$-dimensional random variable $\bm x_0 \sim \mathbb R^D$ with an unknown distribution $q_0(\bm x_0)$ \footnote{To ensure clarity, we establish the convention that the subscript $t$ denotes the timestep in the diffusion process, while the superscript $t$ represents the timestep in sequential decision-making problem.}. DMs gradually transform samples from a simple distribution $q_T(\bm x_T)$ into samples from $q_0(\bm x_0)$ \citep{kingma2021variational, ho2020ddpm}, which is accomplished by solving a reverse Stochastic Differential Equation (SDE) or Ordinary Differential Equation (ODE) \citep{song2021scorebased}:
\begin{align}
    \dd\bm x_t&=[f(t)\bm x_t-g^2(t)\nabla_{\bm x}\log q_t(\bm x_t)]\dd t + g(t)\dd \bar{\bm w}_t,~\bm x_T\sim q_T(\bm x_T),\label{eq:reverse_sde} \\
    \dd\bm x_t&=[f(t)\bm x_t-\frac{1}{2}g^2(t)\nabla_{\bm x}\log q_t(\bm x_t)]\dd t, ~\bm x_T\sim q_T(\bm x_T),\label{eq:reverse_ode}
\end{align}
 where $\bm\bar{w}_t$ is a standard Wiener process in the reverse time, $f(t)=\frac{\dd\log\alpha_t}{\dd t},~ g^2(t)=\frac{\dd\sigma^2_t}{\dd t}-2\sigma^2_t\frac{\dd\log\alpha_t}{\dd t}$, and $\bm x_t=\alpha_t\bm x_0 + \sigma_t\bm\epsilon,~ \bm\epsilon\sim\mathcal{N}(\bm 0, \bm I)$. The \textit{noise schedule} $\alpha_t, \sigma_t \in \mathbb R^+$ are differentiable functions of $t$ such that the \textit{signal-to-noise-ratio} (SNR) $\alpha_t^2/\sigma_t^2$ is strictly decreasing w.r.t $t$. The training of DMs involves using a neural network parameterized by $\theta$ to estimate the unknown term within the SDE or ODE. Different DMs may incorporate varying parameterizations. For instance, diffusion SDE uses a network to estimate a scaled \textit{score function} $\bm\epsilon_\theta(\bm x_t, t)\approx -\sigma_t\nabla_{\bm x}\log q_t(\bm x_t)$ \citep{ho2020ddpm, song2021ddim, lu2022dpmsolver, song2021scorebased}, while EDM estimates clean data $\bm D_\theta(\bm x_t, t)\approx(\bm x_t - \sigma_t^2\nabla_{\bm x}\log q_t(\bm x_t))/\alpha_t$ \citep{karras2022edm}. The sampling process of DMs involves utilizing numerical solvers to solve the SDE or ODE. DDPM \citep{ho2020ddpm} and DDIM \citep{song2021ddim} solve the first-order discretization of \Cref{eq:reverse_sde} and \Cref{eq:reverse_ode}. DPM-Solver \citep{lu2022dpmsolver, lu2023dpmsolverpp} leverages the semi-linearity of the reverse ODE in \Cref{eq:reverse_ode} for exact solutions, eliminating errors in the linear terms, resulting in a higher sample quality. EDM \citep{karras2022edm} uses a specially designed score function preconditioning and $2^{\text{nd}}$-order Heun's method to solve the reverse ODE, also improving the sample quality. Understanding training and sampling as separate processes enables the seamless selection of varying sampling steps and solvers during the generation process without additional training. Some other SDE/ODE-based generative models, such as Rectified Flow \citep{liu2023rectifiedflow}, can also be understood through this lens by using a network to estimate the unknown drift force $\bm v_\theta(\bm x_t, t)\approx(\bm x_0-\bm x_T)$ in a straight ODE $\dd\bm x_t=\bm v_\theta(\bm x_t, t)\dd t$ and solving it by Euler solver. See \Cref{append:foundation_diffusion} for more details.

\section{Revisiting Diffusion Models in Decision Making Scenarios}

\begin{figure}[t]
\centering
\includegraphics[width=1.0\textwidth]{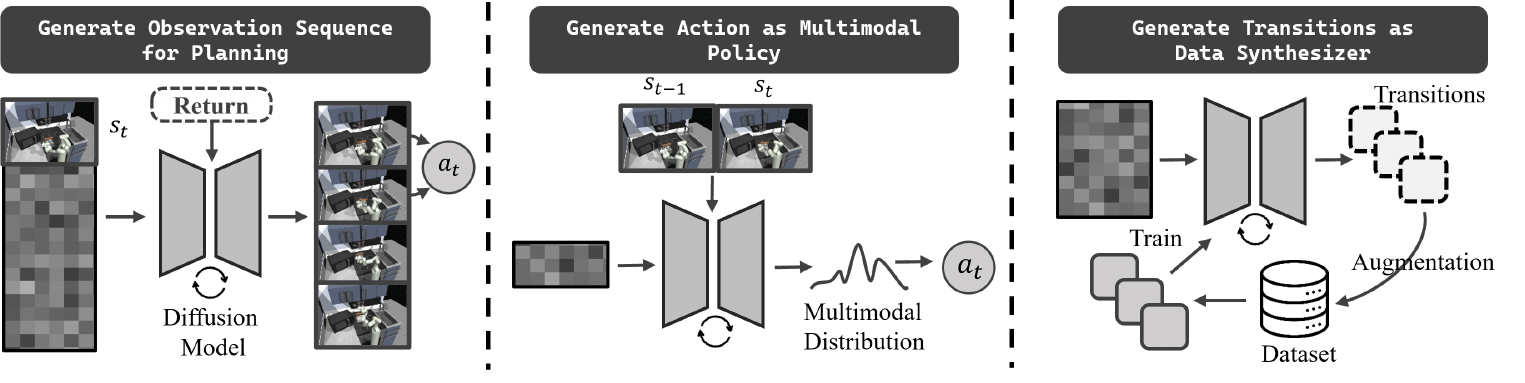}
\vspace{-15pt}
\caption{\small{\textbf{Diffusion Models Mainly Play Three Roles in Decision-Making Scenarios.}} Planner~\cite{janner2022planning}: Acting as planners to make better decisions from a long-term perspective. Policy~\cite{pearce2023imitating}: Serving as policies to support complex multimodal-distribution modeling. Data Synthesizer~\cite{lu2024synthetic}: Performing data augmentation to assist model training.
}
\vspace{-12pt}
\label{fig:role}
\end{figure}


As shown in \Cref{fig:role}, current works applying DMs on decision-making mainly fall into three categories~\citep{zhu2023diffusion}: \textit{generating long-term trajectories and executing like \textbf{planners}}, \textit{replacing the conventional Gaussian policies with multimodal diffusion \textbf{policies}} and \textit{serving as \textbf{data synthesizers} to assist model training}. This section briefly introduces each category, outlines the technical module-design requirements, and summarizes the challenges of designing a general framework.

\textbf{Planner.} Planning 
refers to generating trajectories $\bm x$, which can be either sequence of states or state-action pairs, to maximize the cumulative reward and selecting actions to track the trajectory \citep{hafner2019planet, hansen2022tdmpc, hansen2024tdmpc2}. DMs can simultaneously generate super-long, high-quality trajectories, preventing severe compounding errors occurred in previous planning algorithms \citep{janner2022planning, dong2024diffuserlite}. Assume the trajectory starts at $t=\tau$ and ends at $\mathcal T$, diffusion planner sample from an optimality-conditioned trajectory distribution $p(\bm x|\mathcal O^{\tau:\mathcal T})$ \citep{janner2022planning} or a reward-conditioned distribution $p(\bm x|\sum_{t=\tau}^{\mathcal{T}}r^t)$ \citep{ajay2022conditional}. At each inference step, diffusion planner generates a set of candidate trajectories $\{\bm x_0\}$, selects the local optimal $\bm x_0^*$, and then extracts the action to execute. Typically, these algorithms freeze certain known parts of the trajectories during the diffusion process, such as history trajectories, current states, and future goals, turning the generation into an inpainting problem  \citep{janner2022planning, ajay2022conditional, dong2024diffuserlite, hu2023instructed}. This feature necessitates the demand for a flexible masking mechanism to design frozen parts and freely alter the planning properties.


\textbf{Policy.}
Policy is typically a state-conditioned action distribution $\pi_\theta(\bm a|\bm s)$. DMs' strong distribution modeling capability allows them to effectively replace commonly used deterministic or Gaussian policies \citep{kumar2020cql, kostrikov2022iql, fujimoto2021td3bc} in both RL and IL settings. In RL settings, researchers have explored incorporating diffusion policies as actors in actor-critic frameworks \citep{wang2023diffusion, kang2024efficient}, as well as directly fitting the optimal policy derived from generalized constrained policy search (CPS) \citep{hansen2023idql, chen2023sfbc}. 
These works focus on the combination of DMs and RL components, where RL may guide the generation \citep{lu2023contrastive}, evaluate action selection \citep{chen2023sfbc, hansen2023idql}, or even influence DM training \citep{wang2023diffusion}. In IL settings, researchers focus more on complex network designs to support effective guided sampling \citep{chi2023diffusion, pearce2023imitating, Ze2024DP3, octo_2023}, which processes rich-modality agent perception, including low-dim physical quantities~\cite{pearce2023imitating}, RGB images~\cite{chi2023diffusion}, 3D point clouds~\cite{Ze2024DP3}, and even language instructions~\cite{zhang2022language}. A separated guided sampling module can help researchers divide and conquer, avoiding engineering difficulties caused by coupled structures.

\textbf{Data Synthesizer.} Utilizing synthetic data, which can be either transitions or trajectories, from generative models to assist policy learning has been proven effective \citep{baris2021investigation,clio2022s2p}. Introducing DMs as the generative backbone promotes synthetic quality \citep{lu2024synthetic}, addressing the lack of fidelity in previous works. Unlike Planner or Policy, Data Synthesizer does not directly engage in decision-making and, therefore, requires a flexible and modular library compatible with different DM usage paradigms. 

In summary, building a general modular DM library for decision-making should meet the following criteria: (1) Implement decoupled modules for DM backbones and network architectures to ensure compatibility with different roles. 
(2) Incorporate decision-making specific features into module design, e.g., masking and advanced sampling mechanisms.
(3) Develop an algorithmic pipeline that seamlessly integrates the modules and mechanisms, catering to different DM usage paradigms.

\section{CleanDiffuser}

\subsection{Overview}
\vspace{-3pt}

Based on the analysis above, we illustrate the core sub-modules in \Cref{fig:framework} and summarize them as follows: (1) \texttt{Diffusion Models.} Existing works~\cite{chi2023diffusion, janner2022planning} often tightly couple SDE/ODE, solvers, and algorithm-specific components in their code implementations, making it challenging for practitioners to read and modify. \alg aims to decouple diffusion models as an external module, with internally independent core parts for SDE/ODE and solvers. This design allows users to freely change between solvers and adjust sampling steps with no cost after training. (2) \texttt{Network Architectures} play a crucial role in diffusion-based decision-making algorithms, influencing generative characteristics and indirectly altering algorithm mechanisms \citep{dong2024diffuserlite, lu2024synthetic}. Currently, there is no single architecture that has emerged as the best choice for all scenarios. Therefore, in this module, \alg aims to implement the most commonly used architectures to date, leaving ample room for customization and exploration. (3) \texttt{Guided Sampling Methods.} Existing works employ a rich guided sampling design, ranging from scalar \citep{janner2022planning, ajay2022conditional} to complex multi-modal environment perception \citep{chi2023diffusion, pearce2023imitating}. However, their code implementations often couple guided sampling with other components, making independent guidance design challenging. \alg aims to decouple this aspect as a separate module, providing users with ample customization space. (4) \texttt{Environment Interface \& Dataloader.} \alg provides a consistent environment interface and efficient dataloader for easy usage and evaluation of policy performance.

\subsection{Modular Design}

\begin{figure}[tb]
\centering
\includegraphics[width=0.9\textwidth]{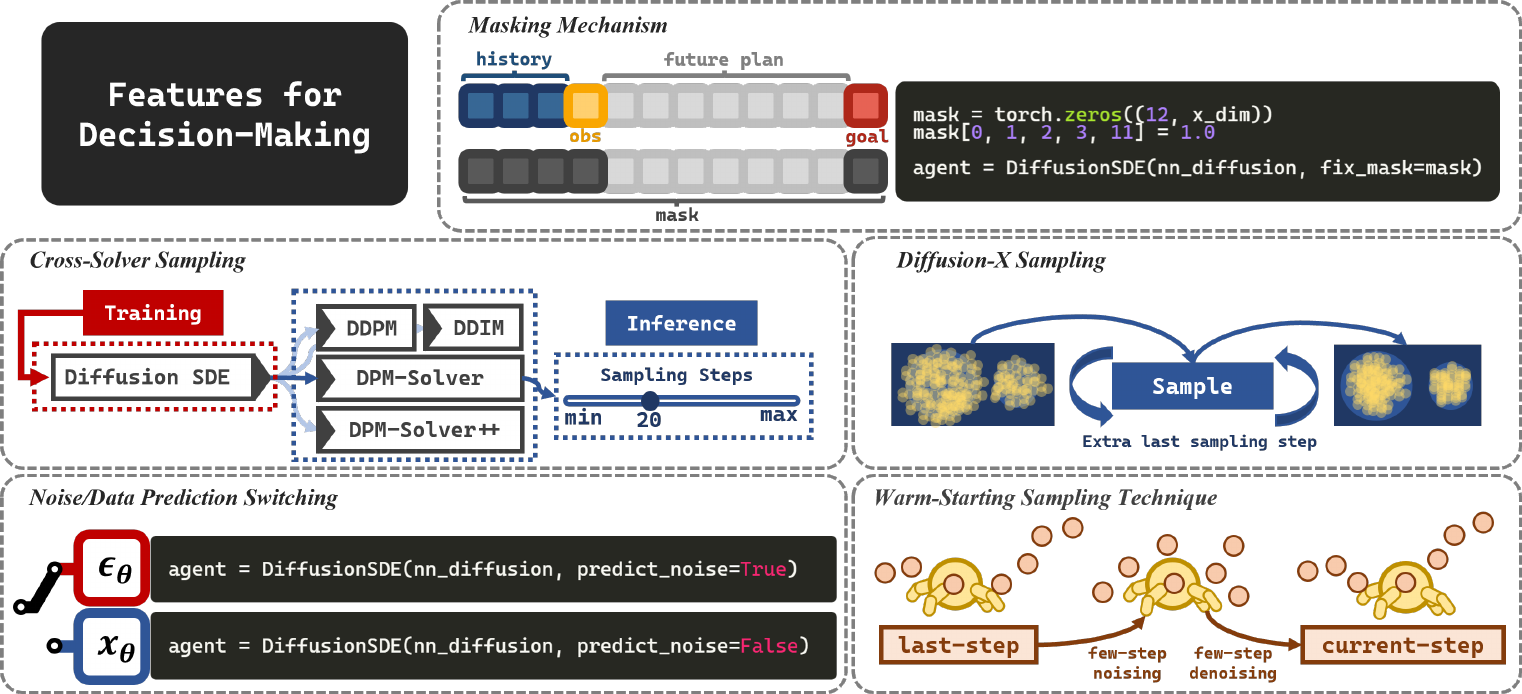}
\caption{\small{\textbf{Features of \alg Designed for Decision-Making Introduced in \Cref{sec:features}.}}}
\label{fig:features}
\vspace{-10pt}
\end{figure}



\textbf{Advanced Diffusion Models Support.} \label{sec:features}
\alg supports advanced diffusion models such as DDPM \citep{ho2020ddpm}, DDIM \citep{song2021ddim}, DPM-Solver \citep{lu2022dpmsolver}, DPM-Solver++ \citep{lu2023dpmsolverpp}, EDM \citep{karras2022edm}, and Rectified Flow \citep{liu2023rectifiedflow}, which share a unified API calling, see \Cref{append:implemented_dms}. Our implementation features the following:
\begin{itemize}[leftmargin=*,itemsep=0pt,topsep=0pt]
    \item \textit{Masking Mechanism.} DM-based decision-making algorithms may incorporate masks to freeze certain known parts and alter the use of generated data \citep{janner2022planning, dong2024diffuserlite}. For example, as demonstrated in \Cref{fig:features} (top), during trajectory generation, one may use a history trajectory as context, retain the current state to provide instant information, and supply a goal to steer the trajectory towards it. The masking mechanism provides a simple interface, using a binary vector describing the freeze requirements. All additional computational processing due to masking is handled internally in the code so that users can concentrate on designing other components.
    \item \textit{Cross-Solver Sampling.} DMs in \alg are implemented with two core parts: \textit{SDE/ODE} and \textit{solver}. Training involves using neural networks to fit the parameterized terms in the SDE/ODE, e.g., the score function in diffusion SDE, and is unrelated to the solvers. This design allows one trained diffusion model to choose varying sampling steps and different solvers during generation without additional cost. For example, after training a decision-making algorithm based on diffusion SDE, one can seamlessly use varying sampling steps and switch between DDPM, DDIM, DPM-Solver, and DPM-Solver++ during inference, greatly facilitating researchers conducting ablation studies and analyses across different diffusion backbones.
    \item \textit{Diffusion-X Sampling.} Considering the significant negative impact of out-of-distribution (OOD) samples in decision-making tasks, the Diffusion-X sampling process is proposed to include additional repeating denoising steps at the last sampling step \citep{pearce2023imitating}. This approach helps concentrate the generated samples in high-likelihood regions, reducing OOD issues.
    \item \textit{Noise/Data Prediction Switching.} Neural networks in DMs can be utilized for predicting noise as well as clean data. In decision-making tasks, the former simplifies optimization by avoiding the direct generation of complex data samples \citep{ajay2022conditional, ho2020ddpm, wang2023diffusion}, while the latter can introduce thresholding methods to constrain samples and prevent OOD generation \citep{janner2022planning, lu2023dpmsolverpp, kang2024efficient}. Existing methods lack a systematic exploration of the effects resulting from these two parameterization approaches. \alg implements noise/data prediction as a switch, depicted in \Cref{fig:features}, to offer researchers a flexible and convenient way to compare between the two approaches.
    \item \textit{Warm-Starting Sampling Technique.} Decision-making dynamics exhibit a certain consistency over time, implying that samples generated at adjacent decision-time steps have similarities. Inspired by this, the warm-starting sampling technique proposes adding a small amount of noise to the samples generated at the previous time step and then conducting a few denoising steps to generate samples of sufficient quality for the current time step. This trick can trade off a small amount of accuracy for an increase in decision frequency and can be useful in real-world applications.
\end{itemize}



\begin{figure}[tb]
\vspace{-12pt}
\centering
\includegraphics[width=0.9\textwidth]{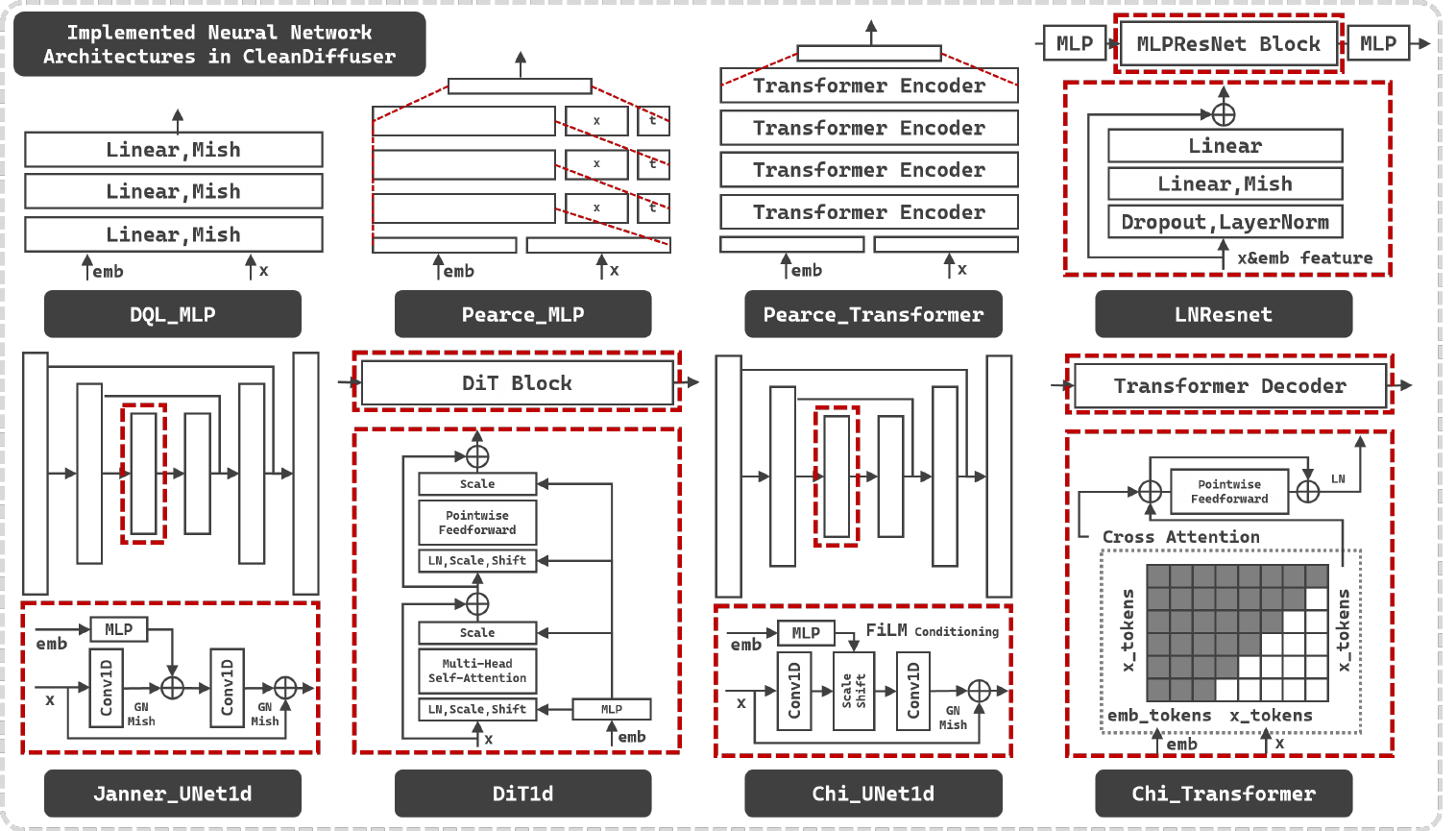}
\vspace{-5pt}
\caption{\small{\textbf{Visualization of Implemented Network Architectures in \alg.}}}
\label{fig:nn}
\vspace{-20pt}
\end{figure}

\textbf{Network Architectures Designed for Decision-Making.} \alg incorporates 8 popular network architectures designed for decision-making, as demonstrated in \Cref{fig:nn}, including:
\begin{itemize}[leftmargin=*,itemsep=0pt,topsep=0pt]

\item \texttt{DQL\_MLP} \citep{wang2023diffusion} is a simple yet efficient MLP architecture for action generation proposed in DQL.

\item \texttt{LNResnet} \citep{hansen2023idql} is a residual MLP with Dropout and LayerNorm to enhance action quality.

\item \texttt{Pearce\_MLP} \citep{pearce2023imitating}, referred to as MLPSieve in DiffusionBC paper, is a residual MLP, which concatenates original inputs to each hidden feature.

\item \texttt{Janner\_UNet1d} \citep{janner2022planning} inherits from the classic image-generation network architecture used in DDPM++ and NCSN++ \citep{song2021scorebased}, and is modified for trajectory generation. This architecture can generate variable-length trajectories \citep{janner2022planning}, which enhances inference flexibility.

\item \texttt{Chi\_UNet1d} \citep{chi2023diffusion} incorporates FiLM conditioning \citep{perez2018film} in \texttt{Janner\_UNet1d} to enhance the reception of sequential observation conditions, achieving excellent performance in IL tasks.

\item \texttt{DiT1d} \citep{dong2023aligndiff} inherits from the transformer DM network backbone \citep{peebles2023dit} and is modified for trajectory generation, showing better training stability and sample quality compared to \texttt{Janner\_UNet1d}.

\item \texttt{Pearce\_Transformer} \citep{pearce2023imitating} replaces the structure in \texttt{Pearce\_MLP} with the multi-head self-attention, which sacrifices efficiency for better action generation quality.

\item \texttt{Chi\_Transformer} \citep{chi2023diffusion} employs a transformer decoder architecture and a special cross-attention mask to enhance the reception of conditions, achieving performance similar to \texttt{Chi\_UNet1d}.

\end{itemize}

These network architectures have been proven effective for decision-making tasks in previous works and widely referenced or directly applied in other algorithms \citep{chen2023sfbc, dong2024diffuserlite, liang2023adaptdiffuser, ni2023metadiffuser, li2023hdmi, he2024diffcps}. In \alg, all these architectures inherit from the same parent class and share a standard API calling, making it easy for researchers to design new architectures based on the foundations.

\textbf{Guided Sampling.} Two guided sampling methods, CG \citep{dhariwal2021classifierguidance} and CFG \citep{ho2021classifierfree}, are presented in the form of \textit{Classifier} and \textit{Condition Network}, which are completely decoupled from the DM network architecture. Users can focus solely on processing condition information without worrying about the interaction with DMs and eventually integrate them with DMs in a switch-like manner.

\textbf{Environment Interface and Efficient Dataloader:} To facilitate benchmark evaluation, we encapsulate Gym-like~\cite{brockman2016openai} API for all environments, implementing visualization, multi-step interaction, and parallel sampling through various wrappers. This makes it convenient for researchers to reuse and extend. Additionally, we implement efficient I/O based on Zarr~\cite{zarr} library for large-scale datasets and combine it with PyTorch's DataLoader~\cite{pytorch} for batch data processing and training, which allows for flexible data access even with limited memory. \alg also provides Wandb~\cite{wandb} logging support and Hydra~\cite{Yadan2019Hydra} configuration to facilitate experiment tracking. We provide YAML configuration files for each experiment, ensuring full reproducibility without tuning hyperparameters.

\subsection{From Decoupled Modules to Integrated Pipelines} 

\begin{figure}[bht]
\centering
\includegraphics[width=1\textwidth]{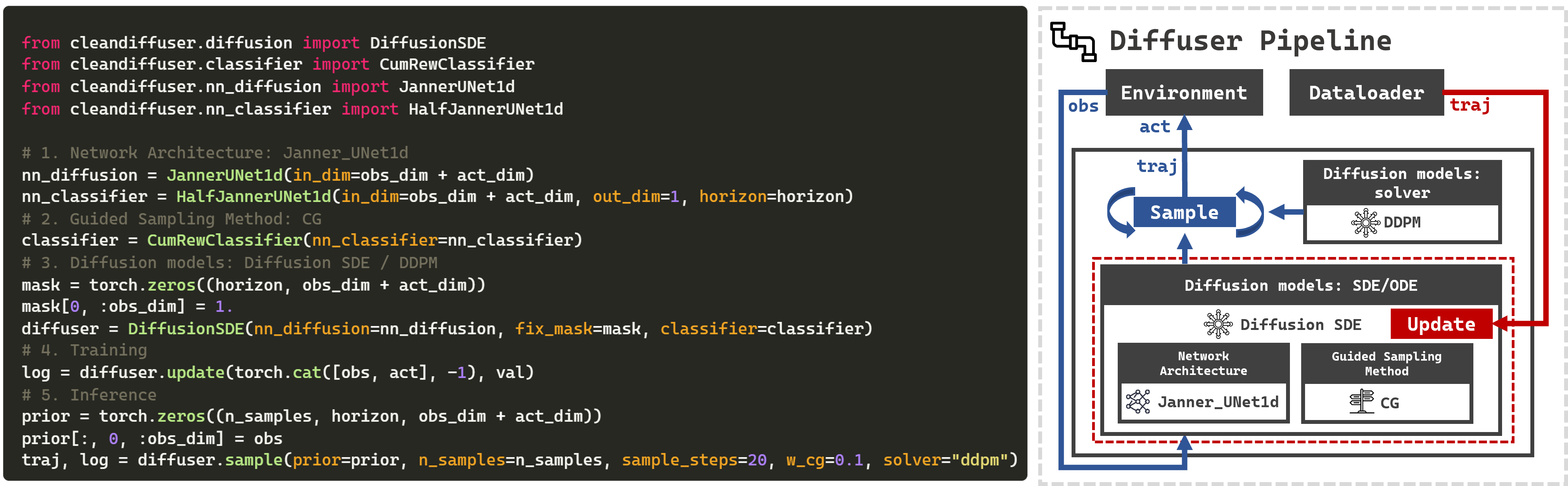}
\vspace{-8pt}
\caption{\small{\textbf{Diffuser Implementation with \alg.} The left part is a minimal code example showcasing simplicity and readability, and the right part provides a code explanation where the algorithm implementation can be entirely represented as a combination of building blocks, showing an example of various pipelines.}}
\label{fig:pipeline}
\end{figure}

With \alg, developing algorithms can be much more straightforward because users only need to select the desired building blocks and assemble them into a pipeline. As shown in \Cref{fig:pipeline}, a Diffuser implementation example that uses \texttt{Janner\_UNet1d} as the network architecture for generating trajectories, employs a \textit{Classifier} for guided sampling to maximize the cumulative reward of generated trajectories, selects Diffusion SDE as the diffusion backbone, and performs sampling using DDPM. Assembling these modules constructs a pipeline, a simple yet efficient Diffuser implementation. In this way, users can easily understand the differences and properties of algorithms and adjust them by simply replacing the building blocks. In \alg, we implement various diffusion-based decision-making algorithms in this module-to-pipeline style, offering a diverse set of examples for practitioners to implement their applications with \alg. The implemented algorithms include three diffusion planners: Diffuser \citep{janner2022planning}, Decision Diffuser (DD) \citep{ajay2022conditional}, and AdaptDiffuser \citep{liang2023adaptdiffuser}; five diffusion policies: DiffusionPolicy \citep{chi2023diffusion}, DiffusionBC \citep{pearce2023imitating}, DQL \citep{wang2023diffusion}, EDP \citep{kang2024efficient}, and IDQL \citep{hansen2023idql}; one diffusion data synthesizer: SynthER \citep{lu2024synthetic}. See \Cref{append:algos} for details.

\section{Experiments}

Due to space limitations in the main text, we introduce details of all benchmarks and datasets used in our experiments in \Cref{append:env_details}, and present additional experiments in \Cref{append:additional_exp}.

\vspace{-2pt}
\subsection{Offline Reinforcement Learning}
\begin{table}[htb]
\centering
\caption{\small{\textbf{Evaluation Results of Offline RL Benchmark.} The performance of diffusion-based offline RL algorithms implemented by \alg on the D4RL benchmark \citep{fu2020d4rl}. Results correspond to the mean and standard error over 150 episode seeds; the highest scores are emphasized in bold.}}
\vspace{-2pt}
\label{tab:rl_table}
\scalebox{0.6}{
\begin{tabular}{llcccccccc}
\toprule
\textbf{Dataset} & \textbf{Environment} & \textbf{BC} & \textbf{SynthER} & \textbf{Diffuser} & \textbf{DD} & \textbf{AdaptDiffuser} & \textbf{DQL} & \textbf{EDP} & \textbf{IDQL} \\ \midrule
\multirow{3}{*}{Medium-Expert} & HalfCheetah & $55.2$ & $94.8\pm0.0$ & $90.3\pm0.1$ & $88.9\pm1.9$ & $90.4\pm0.1$ & $95.5\pm0.1$ & $\bm{95.8\pm0.1}$ & $91.3\pm0.6$ \\
 & Hopper & $52.5$ & $76.6\pm0.4$ & $107.2\pm0.9$ & $110.4\pm0.6$ & $109.3\pm0.3$ & $\bm{111.1\pm0.4}$ & $110.8\pm0.4$ & $110.1\pm0.7$ \\
 & Walker2d & $107.5$ & $110.0\pm0.0$ & $107.4\pm0.1$ & $108.4\pm0.1$ & $107.7\pm0.1$ & $\bm{111.6\pm0.0}$ & $110.4\pm0.0$ & $110.6\pm0.0$ \\ \midrule
\multirow{3}{*}{Medium} & HalfCheetah & $42.6$ & $48.3\pm0.0$ & $43.8\pm0.1$ & $45.3\pm0.3$ & $44.3\pm0.2$ & $\bm{52.3\pm0.2}$ & $50.8\pm0.0$ & $51.5\pm0.1$ \\
 & Hopper & $52.9$ & $51.9\pm0.1$ & $89.5\pm0.7$ & $\bm{98.2\pm0.1}$ & $95.5\pm1.1$ & $96.5\pm1.3$ & $72.6\pm0.2$ & $70.1\pm2.0$ \\
 & Walker2d & $75.3$ & $86.6\pm0.0$ & $79.4\pm1.0$ & $79.6\pm0.9$ & $83.8\pm1.1$ & $86.8\pm0.0$ & $86.5\pm0.2$ & $\bm{88.1\pm0.4}$ \\ \midrule
\multirow{3}{*}{Medium-Replay} & HalfCheetah & $36.6$ & $43.4\pm0.0$ & $36.0\pm0.7$ & $42.9\pm0.1$ & $36.7\pm0.8$ & $\bm{47.9\pm0.0}$ & $44.9\pm0.4$ & $46.5\pm0.3$ \\
 & Hopper & $18.1$ & $24.7\pm0.1$ & $91.8\pm0.5$ & $99.2\pm0.2$ & $91.2\pm0.1$ & $\bm{101.6\pm0.0}$ & $83.0\pm1.7$ & $99.4\pm0.1$ \\
 & Walker2d & $26.0$ & $88.6\pm0.4$ & $58.3\pm1.8$ & $75.6\pm0.6$ & $82.9\pm1.5$ & $\bm{98.2\pm0.1}$ & $87.0\pm2.6$ & $89.1\pm2.4$ \\ \midrule
\multicolumn{2}{c}{\textbf{Average}} & $51.9$ & $69.4$ & $78.2$ & $83.2$ & $82.4$ & $\bm{89.0}$ & $82.4$ & $84.1$ \\ \midrule
Mixed & Kitchen & $51.5$ & $0.0\pm0.0$ & $52.5\pm2.5$ & $\bm{75.0\pm0.0}$ & $51.8\pm0.8$ & $62.5\pm1.5$ & $50.2\pm1.8$ & $66.5\pm4.1$ \\
Partial & Kitchen & $38.0$ & $0.0\pm0.0$ & $55.7\pm1.3$ & $56.5\pm5.8$ & $55.5\pm0.4$ & $63.5\pm1.8$ & $40.8\pm1.5$ & $\bm{66.7\pm2.5}$ \\ \midrule
\multicolumn{2}{c}{\textbf{Average}} & $44.8$ & $0.0$ & $54.1$ & $65.8$ & $53.7$ & $63.0$ & $45.5$ & $\bm{66.6}$ \\ \midrule
\multirow{2}{*}{Play} & Antmaze-Medium & $0.0$ & $0.0\pm0.0$ & $6.7\pm5.7$ & $8.0\pm4.3$ & $12.0\pm7.5$ & $\bm{86.0\pm1.8}$ & $73.3\pm6.2$ & $67.3\pm5.7$ \\
 & Antmaze-Large & $0.0$ & $0.0\pm0.0$ & $17.3\pm1.9$ & $0.0\pm0.0$ & $5.3\pm3.4$ & $\bm{83.3\pm2.5}$ & $33.3\pm1.9$ & $48.7\pm4.7$ \\
\multirow{2}{*}{Diverse} & Antmaze-Medium & $0.8$ & $0.0\pm0.0$ & $2.0\pm1.6$ & $4.0\pm2.8$ & $6.0\pm3.3$ & $\bm{94.7\pm2.5}$ & $52.7\pm1.9$ & $83.3\pm5.0$ \\
 & Antmaze-Large & $0.0$ & $0.0\pm0.0$ & $27.3\pm2.4$ & $0.0\pm0.0$ & $8.7\pm2.5$ & $\bm{61.3\pm8.4}$ & $41.3\pm3.4$ & $40.0\pm11.4$ \\ \midrule
\multicolumn{2}{c}{\textbf{Average}} & $0.2$ & $0.0$ & $13.3$ & $3.0$ & $8.0$ & $\bm{81.3}$ & $50.2$ & $59.8$ \\ \bottomrule
\end{tabular}}
\end{table}

\textbf{Setup.} We evaluate 7 diffusion-based offline RL algorithms with \alg, including SynthER, Diffuser, DD, AdaptDiffuser, DQL, EDP, and IDQL, on 15 tasks in the D4RL \citep{fu2020d4rl}, covering locomotion, manipulation, and navigation. We reuse the hyperparameters of the original paper as possible and give the full hyperparameters in \Cref{append:hyperparams}. The results are presented in \Cref{tab:rl_table}.

\begin{table}[tb]
\centering
\small
\caption{\small{\textbf{Evaluation Results of Offline IL Benchmark.} The metrics show success rate for Robomimic and Relay-Kitchen, target area coverage for PushT. We report mean performance of last checkpoint denoted as \textsc{Last} and max performance of the last 10 checkpoints (3 for image tasks) denoted as \textsc{Max}, with each averaged over 3 seeds and 50 episodes. We show the performance of (\textsc{Last} $/$ \textsc{Max}). $^*$The results are obtained from the \cite{chi2023diffusion}.}} 
\label{tab:il_table}
\vspace{-3pt}
\scalebox{0.7}{
\setlength{\tabcolsep}{14pt}
\begin{tabular}{@{}lccccccc@{}}
\toprule
\multirow{2}{*}{\textbf{Task Name}} &
  \multicolumn{1}{l}{\multirow{2}{*}{\textbf{LSTM-GMM$^*$}}} &
  \multirow{2}{*}{\textbf{ACT}} &
  \multicolumn{3}{c}{\textbf{DiffusionPolicy}} &
  \multicolumn{2}{c}{\textbf{DiffusionBC}} \\ \cmidrule(l){4-8} 
 &
  \multicolumn{1}{l}{} &
   &
  \textbf{\texttt{DiT1d}} &
  \textbf{\texttt{Chi\_UNet1d}} &
  \textbf{\texttt{Chi\_TFM}} &
  \textbf{\texttt{DiT1d}} &
  \textbf{\texttt{Pearce\_MLP}} \\ \midrule
\textit{\textbf{Low dim}} & \multicolumn{1}{l}{} & \multicolumn{1}{l}{} &                    &                    &                    &                    &            \\ \midrule
pusht                     & 0.59/0.70            &   0.99/\textbf{1.00}        & \textbf{1.00/1.00} & 0.99/\textbf{1.00}          & 0.94/\textbf{1.00}          & 0.99/0.99          & 0.99/0.99  \\
pusht-keypoints           & 0.61/0.67            &     0.99/\textbf{1.00}         & 0.99/\textbf{1.00}          & \textbf{1.00/1.00} & 0.99/0.99          & \textbf{1.00/1.00} & 0.99/0.99  \\
relay-kitchen             & 0.75/0.79            &    0.72/0.76        & \textbf{1.00/1.00} & 0.99/\textbf{1.00}          & 0.99/0.99          & 0.67/0.81          & 0.81/0.89  \\
lift-ph                   & 0.96/\textbf{1.00}            & 0.98/\textbf{1.00}            & \textbf{1.00/1.00} & \textbf{1.00/1.00} & \textbf{1.00/1.00} & \textbf{1.00/1.00} & 0.99/\textbf{1.00}  \\
lift-mh                   & 0.93/\textbf{1.00}            & 0.98/\textbf{1.00}            & \textbf{1.00/1.00} & \textbf{1.00/1.00} & \textbf{1.00/1.00} & 0.99/\textbf{1.00}          & 0.92/\textbf{1.00}  \\
can-ph                    & 0.91/\textbf{1.00}            & 0.92/0.98            & \textbf{1.00/1.00} & 0.99/\textbf{1.00} & 0.99/\textbf{1.00} & 0.99/\textbf{1.00}          & 0.91/\textbf{1.00}  \\
can-mh                    & 0.81/\textbf{1.00}            & 0.90/0.98            & 0.95/0.98          & \textbf{0.99/1.00} & 0.91/\textbf{1.00}          & 0.91/0.98          & 0.77/0.88  \\
square-ph                 & 0.73/0.95            & 0.80/0.90            & 0.85/0.96          & \textbf{0.93/0.98} & 0.87/0.96          & 0.68/0.76          & 0.66/0.76  \\
square-mh                 & 0.59/0.86            & 0.46/0.72            & 0.58/0.74          & \textbf{0.87/0.96} & 0.67/0.86          & 0.50/0.68          & 0.42/0.52  \\
transport-ph              & 0.47/0.76            & 0.64/0.85            & 0.47/0.64          & \textbf{0.79/0.92} & 0.67/0.84          & 0.35/0.54          & 0.17/0.34  \\
transport-mh              & 0.20/0.62            & 0.40/0.68            & 0.25/0.44          & \textbf{0.58/0.72} & 0.23/0.52          & 0.14/0.28          & 0.00/0.04  \\
toolhang-ph               & 0.31/0.67            & 0.64/0.82            & 0.38/0.58          & 0.72/0.90          & \textbf{0.90/0.96} & 0.49/0.66          & 0.15/0.36  \\ \midrule
\textbf{Average}          & 0.66/0.84            &    0.79/0.89         & 0.79/0.86          & \textbf{0.90/0.96}          & 0.85/0.93          & 0.73/0.81          & 0.65/0.73  \\ \midrule
\textit{\textbf{Image}} &
  \multicolumn{1}{l}{} &
  \multicolumn{1}{l}{} &
  \multicolumn{1}{l}{} &
  \multicolumn{1}{l}{} &
  \multicolumn{1}{l}{} &
  \multicolumn{1}{l}{} &
  \multicolumn{1}{l}{} \\ \midrule
pusht-image               & 0.54/0.69                 &   0.99/\textbf{1.00}          & 0.99/\textbf{1.00}               & \textbf{1.00/1.00}               & 0.98/0.99               & 0.10/0.19               & 0.53/0.64       \\
lift-ph                   & 0.96/\textbf{1.00}                 &  \textbf{1.00/1.00}                 & \textbf{1.00/1.00}         & \textbf{1.00/1.00}         & \textbf{1.00/1.00}         & \textbf{1.00/1.00}         & 0.94/0.98 \\
lift-mh                   & 0.95/\textbf{1.00}               &     \textbf{1.00/1.00}           & \textbf{1.00/1.00}         & \textbf{1.00/1.00}         & 0.99/\textbf{1.00}         & 0.88/\textbf{1.00}         & 0.94/0.98 \\
can-ph                    & 0.88/\textbf{1.00}                 & 0.98/0.98                 & 0.97/\textbf{1.00}               & \textbf{0.99/1.00}               & 0.98/\textbf{1.00}               & 0.92/0.94               & 0.89/0.94       \\
can-mh                    & 0.90/\textbf{0.98}                 & 0.94/0.94                  & 0.90/0.92                & \textbf{0.96/0.98}                & 0.89/0.94                & 0.73/0.86                & 0.76/0.84        \\
square-ph                 & 0.59/0.82                 & 0.90/0.90                 & 0.57/0.64               & \textbf{0.95/0.98}               & 0.81/0.86               & 0.21/0.22               & 0.23/0.24       \\
square-mh                 & 0.38/0.64                 &   \textbf{0.84}/0.84               & 0.47/0.68               & 0.83/\textbf{0.94}               & 0.65/0.74               & 0.20/0.30               & 0.15/0.20       \\
transport-ph              & 0.62/0.88                 & 0.79/0.80                 & 0.76/0.84                & 0.88/\textbf{0.96}                & \textbf{0.89/0.96}                & 0.07/0.12                & 0.50/0.66        \\
transport-mh              & 0.24/0.44                 &   0.59/\textbf{0.62}                & 0.52/0.52                & \textbf{0.61/0.62}                & 0.40/0.52                & 0.06/0.08                & 0.10/0.16        \\
toolhang-ph               & 0.49/0.68                 & \textbf{0.69/0.76}                 & 0.59/0.72                & 0.59/0.66                & 0.39/0.44                & 0.06/0.14                & 0.06/0.10        \\ \midrule
\textbf{Average}          & 0.65/0.81                 &   0.87/0.88              & 0.78/0.83                & \textbf{0.88/0.91}                & 0.80/0.85                & 0.42/0.48                & 0.51/0.57        \\ \bottomrule
\end{tabular}%
}
\end{table}
\vspace{-1pt}
\textbf{Key Observation.} \textbf{(O1)} Algorithms reproduced with \alg have achieved, and in some cases exceeded, their official implementations. \textbf{(O2)} Diffusion planners demonstrate no superiority over diffusion policies, especially performing poorly in the Antmaze. Diffusion planners are sensitive to guided sampling and prone to generating OOD trajectories \citep{dong2024diffuserlite}. Enhancing the dynamic legitimacy \citep{ni2023metadiffuser} and introducing the conservative generation \citep{yu2020mopo} may unlock the potential of diffusion planners. \textbf{(O3)} DQL achieves outstanding performance among diffusion policies. Simply incorporating Q-maximizing loss in diffusion training shows stable and surprising performance.

\subsection{Offline Imitation Learning}

\textbf{Setup.} We evaluate DiffusionPolicy and DiffusionBC with different network architectures on 22 tasks across PushT \cite{florence2022implicit}, Relay-Kitchen \cite{gupta2019relay} and Robomimic \cite{robomimic2021} benchmarks. PushT and Robomimic include both low-dim and image-based observations. To validate the imitation capabilities of the DM paradigms, we also compare the RNN-based LSTM-GMM~\cite{robomimic2021} and the Transformer-based ACT~\cite{zhao2023learning} (reproduced). Each method is evaluated with its best-performing action space: position control for DiffusionPolicy and ACT, and velocity control for others. We reuse the hyperparameters of the original paper as much as possible, and key hyperparameters are given in \Cref{append:hyperparams}.

\vspace{-3pt}
\textbf{Key Observation.} \textbf{(O1)} Different network architectures have a significant impact on the performance. Among them, DiffusionPolicy works better than DiffusionBC, and DiffusionPolicy with \texttt{Chi\_UNet1d} has the best performance and training stability~(Performance gap between the best checkpoint and last checkpoint). However, \texttt{Chi\_UNet1d} has large model size and long inference time. We often need to trade-off between inference time and model performance in applications. \textbf{(O2)} Compared to popular RNN or transformer-based imitation learning algorithms, DiffusionPolicy also exhibits stronger performance, but slower inference times due to the multiple network forwards of denoise. We show detailed model size and inference time comparisons and analyses in \cref{app:additional_exp}.

\subsection{Impact of Diffusion Backbones and Sampling Steps}\label{sec:impact_of_db_ss}

\begin{figure}[htb]
\centering
\includegraphics[width=1.0\textwidth]{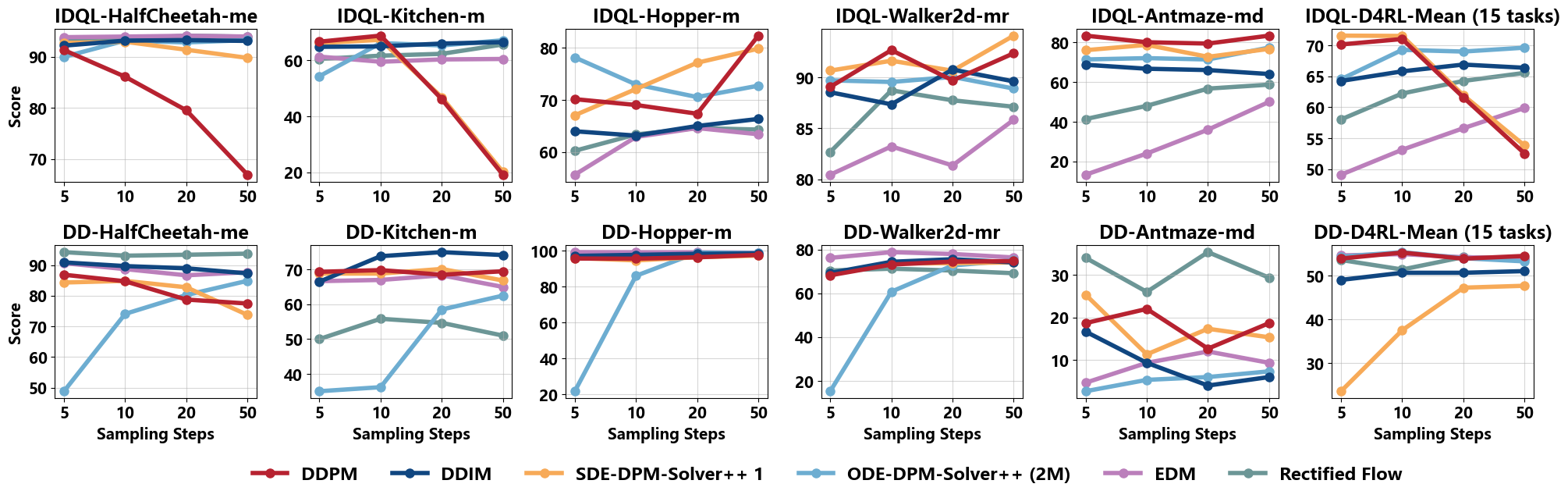}
\caption{\small{\textbf{Impact of Diffusion Backbones and Sampling Steps.} Performance of IDQL and DD with various diffusion backbones and varying sampling steps. Results correspond to the mean over 150 episode seeds.}}
\label{fig:ss_solver_rl}
\end{figure}

\begin{figure}[htb]
\centering
\includegraphics[width=0.9\textwidth]{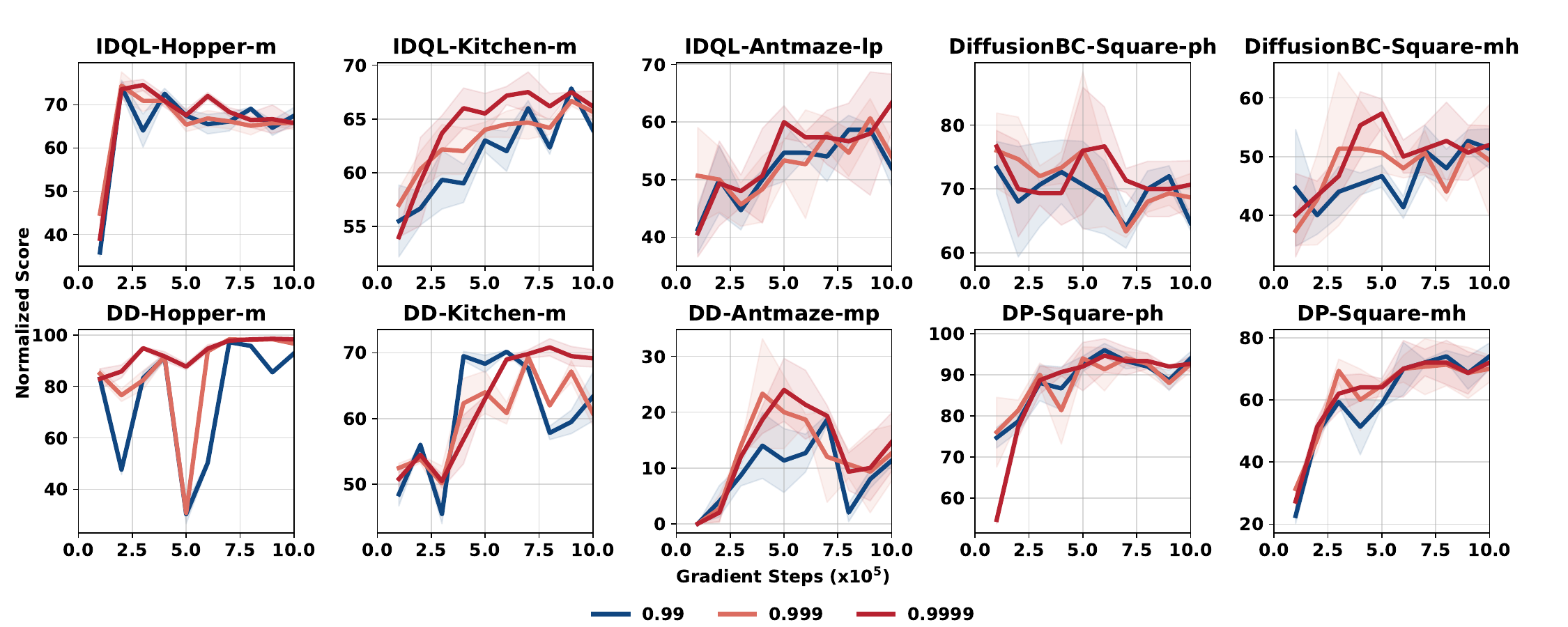}
\caption{\small{\textbf{Impact of EMA Rate and Gradient Steps.} Learning curve of IDQL, DD, DiffusionBC, and DP with varying EMA rates. Results correspond to the mean and standard error over 150 episode seeds.}}
\label{fig:ema}
\end{figure}

Although the impact of diffusion backbones and sampling steps are widely discussed in image generation, little research analyzes them in decision-making. We compare the performance of IDQL and DD, representing policies and planners, respectively, with varying diffusion backbones and sampling steps, showing results on a few tasks in \Cref{fig:ss_solver_rl} and full results in \Cref{append:full_ss_solver}.

\vspace{-3pt}
\textbf{Key Observation.} \textbf{(O1)} An anomaly where performance decreases as the sampling steps increase may happen in some tasks, known as \textit{sampling degradation}. This anomaly has been identified in previous works \citep{kang2024efficient, chen2023sfbc} and remains an open question. Experiments reveal that \textit{sampling degradation} is more likely to occur in medium-expert MuJoCo and Kitchen tasks, possibly due to narrow data distributions. Future research can investigate this issue and offer optimal choices for sampling steps. Additionally, we observe that 5 sampling steps are adequate for most tasks, suggesting that more sampling steps in previous works, e.g., 100 \citep{ajay2022conditional}, are unnecessary. \textbf{(O2)} SDE solvers (DDPM, SDE-DPM-Solver++ 1) perform better in diffusion policies but suffer more from \textit{sampling degradation} than ODE solvers. In diffusion planners, they perform similarly and do not show a \textit{sampling degradation} tendency. While the impact of SDEs and ODEs in image generation has been extensively discussed \citep{nie2024the, lu2023dpmsolverpp}, it remains unexplored in decision-making, suggesting a need for future research.
\textbf{(O3)} High-order solvers (ODE-DPM-Solver++ (2M)) show no superiority over first-order solvers.

\vspace{-3pt}
\subsection{Impact of EMA Rate and Gradient Steps}
\vspace{-3pt}

The exponential moving average (EMA) rate significantly impacts performance \citep{song2021scorebased}. However, limited research has discussed the impact of EMA rate on diffusion-based decision-making algorithms. Previous works tend to use a lower EMA rate, e.g., 0.995 \citep{janner2022planning, ajay2022conditional}, rather than the more common 0.9999 \citep{song2021scorebased, nichol2021improved, liu2023rectifiedflow} used in image generation. We compare the learning curves of IDQL, DD, DiffusionBC, and DiffusionPolicy~(DP) with varying EMA rates and present the results in \Cref{fig:ema}.

\vspace{-3pt}
\textbf{Key Observation.} \textbf{(O1)} A higher EMA rate improves and stabilizes the performance during training, and also helps alleviate \textit{training degradition}, in which model performance drops as the gradient steps increase.\textbf{(O2)} Tested algorithms can almost reach near-convergence performance with around $5\times10^5$ gradient steps even with a high EMA rate. Excessively long gradient steps may be unnecessary.

\section{Conclusion}

We present \alg, the first open-sourced modularized DM library specifically for decision-making algorithms. \alg implements diverse decoupled modules and practical features, supporting different types of DM algorithmic branches. Algorithmic pipelines can be easily implemented by combining sub-modules as simply as building blocks. Extensive experiments validate the library's reliability and versatility, benchmarking the performance of various DM algorithms for future research. We also conduct comprehensive experimental analyses on design choices of DMs, revealing the strengths and challenges of current DM methods.
\alg fills a critical gap in the current landscape by providing a unified library.
We believe \alg lays a solid cornerstone for applying DMs to decision-making tasks and will catalyze further rapid progress in this promising field.
We indicate some limitations, challenges, and future directions in \Cref{append:limit_chall}.

\newpage

\section{Acknowledgements}
This work is supported by the National Natural Science Foundation of China (Grant Nos. 62422605, 92370132).

\bibliography{neurips_2024}
\bibliographystyle{plain}


\newpage
\appendix

\section{Foundation of Diffusion Models}\label{append:foundation_diffusion}

\subsection{SDEs/ODEs and Solvers}
Assume a $D$-dimensional random variable $\bm x_0 \sim \mathbb R^D$ with an unknown distribution $q_0(\bm x_0)$ \footnote{To ensure clarity, we establish the convention that the subscript $t$ denotes the timestep in the diffusion process, while the superscript $t$ represents the timestep in sequential decision-making problem.}. Diffusion Models (DMs) \citep{kingma2021variational, song2021scorebased} define a \textit{forward process} $\{\bm x_t\}_{t\in[0, T]}$ with $T>0$ by the \textit{noise schedule} $\{\alpha_t, \sigma_t\}_{t\in[0,T]}$, such that $\forall t\in[0,T]$, $\bm x_t$ satisfies
\begin{equation}\label{eq:forward}
    \bm x_t = \alpha_t\bm x_0 + \sigma_t \bm\epsilon, ~\bm\epsilon\sim\mathcal{N}(\bm 0, \bm I),
\end{equation}
where $\alpha_t, \sigma_t \in \mathbb R^+$ are differentiable functions of $t$ and the \textit{signal-to-noise-ratio} (SNR) $\alpha_t^2/\sigma_t^2$ is strictly decreasing w.r.t $t$. The forward process in \Cref{eq:forward} can also be described as a stochastic differential equation (SDE) for any $t\in[0, T]$ \citep{kingma2021variational}:
\begin{equation}\label{eq:forward_sde}
    \dd\bm x_t=f(t)\bm x_t\dd t+g(t)\dd \bm w_t, ~\bm x_0 \sim q_0(\bm x_0),
\end{equation}
where $\bm w_t\in\mathbb R^D$ is the standard Wiener process, and $f(t)=\frac{\dd\log\alpha_t}{\dd t}, g^2(t)=\frac{\dd\sigma^2_t}{\dd t}-2\sigma^2_t\frac{\dd\log\alpha_t}{\dd t}$. The SDE forward process in \Cref{eq:forward_sde} has an equivalent \textit{reverse process} from time $T$ to $0$ \citep{song2021scorebased}:
\begin{equation}
    \dd\bm x_t=[f(t)\bm x_t-g^2(t)\nabla_{\bm x}\log q_t(\bm x_t)]\dd t + g(t)\dd \bar{\bm w}_t,~\bm x_T\sim q_T(\bm x_T),
\end{equation}
where $\bm\bar{w}_t$ is a standard Wiener process in the reverse time. One can sample $q_0(x_0)$ by directly solving the SDE in \Cref{eq:reverse_sde}, in which the only unknown term is the \textit{score function} \scorefunc. In practice, a neural network $\bm\epsilon_\theta(\bm x_t)$ parameterized by $\theta$ can be trained to approximate the scaled score function $-\sigma_t\nabla_{\bm x}\log q_t(\bm x_t)$ by minimizing the score matching loss \citep{ho2020ddpm, song2021ddim, song2021scorebased}:
\begin{align}
    \mathcal L(\theta):=&\mathbb E_{t\sim \text{Uniform}(0, T),\bm x_t\sim q_t(\bm x_t)}\left[\Vert\bm\epsilon_\theta(\bm x_t,t)+\sigma_t\nabla_{\bm x}\log q_t(\bm x_t)\Vert^2_2\right] \\
    =&\mathbb E_{t\sim \text{Uniform}(0, T),\bm x_0\sim q_0(\bm x_0),\bm\epsilon\sim\mathcal{N}(\bm 0, \bm I)}\left[\Vert\bm\epsilon_\theta(\bm x_t,t)-\bm\epsilon\Vert^2_2\right].
\end{align}
Since $\bm\epsilon_\theta(\bm x_t, t)$  can be considered as a predicted Gaussian noise added to $\bm x_t$, it is usually called the \textit{noise prediction model}. With a well-trained noise prediction model, SDE in \Cref{eq:reverse_sde} can be solved using numerical solvers, and DDPM \citep{ho2020ddpm} is one such method. However, numerical solvers require discretization from $T$ to $0$, in which the randomness of the Wiener process limits the step size \citep{kloeden2011numerical}. For faster sampling, one can solve the following \textit{probability flow} ODE, which is proven to have the same marginal distribution as that of the SDE for any $t\in[0, T]$ \citep{song2021scorebased}:
\begin{equation}
    \frac{\dd x_t}{\dd t}=f(t)\bm x_t-\frac{1}{2}g^2(t)\nabla_{\bm x}\log q_t(\bm x_t), ~\bm x_T\sim q_T(\bm x_T).
\end{equation}
DDIM \citep{song2021ddim} discretizes the ODE to the first order for solving, achieving almost no loss in quality with fewer sampling steps. DPM-Solver \citep{lu2022dpmsolver, lu2023dpmsolverpp} leverages the semi-linearity of diffusion ODEs in \Cref{eq:reverse_ode} for exact solutions, eliminating errors in the linear terms, resulting in a higher sample quality. Some works also reformulate the framework. EDM \citep{karras2022edm} optimizes the design choices from a perspective of noise schedule and uses a specially designed score function preconditioning to improve the sample quality. Rectified flow \citep{liu2023rectifiedflow}, on the other hand, designs a straight probability flow ODE from the optimal transport (OT) perspective, which can straighten itself through \textit{reflow} procedure. The straight property of Rectified flow allows high-quality generation in very few sampling steps.

\subsection{Guided Sampling Methods}

Guided sampling methods aim to draw samples from $q_0(x_0|y)$ to generate outputs with the characteristics of the label $y$. Depending on whether an additional classifier needs to be trained, guided sampling methods are divided into two categories: classifier guidance (CG) \citep{dhariwal2021classifierguidance} and classifier-free guidance (CFG) \citep{ho2021classifierfree}.

\textbf{Classifier Guidance:}
For conditional sampling, the score function needs to be changed to $\nabla_{\bm x}\log q_t(\bm x_t|\bm y)$, which can be decomposed with the Bayes Theorem:
\begin{equation}\label{eq:bayes}
    \nabla_{\bm x}\log q_t(\bm x_t|\bm y)=\nabla_{\bm x}\log q_t(\bm x_t)+\nabla_{\bm x}\log q_t(\bm y|\bm x_t),
\end{equation}
where the first term can be approximated by the noise prediction model, and the second term is a \textit{noising classifier} that predicts the label $y$ of the corrupt data $\bm x_t$. In practice, an additional neural network $\mathcal C_\phi(\bm x_t,t,\bm y)$ is trained to approximate $\log q_t(\bm y|\bm x_t)$, and its gradient is computed to guide sampling process:
\begin{equation}
\bar{\bm\epsilon}_\theta(\bm x_t, t, \bm y)=\bm\epsilon_\theta(\bm x_t, t) - w\sigma_t\nabla_{\bm x}\mathcal{C}_\phi(\bm x_t,t,\bm y),
\end{equation}
where $w$ stands for the guidance scale. A larger value of $w$ sharpens the classifier, amplifying the influence of the label $y$.

\textbf{Classifier-free Guidance:}
According to \Cref{eq:bayes}, the gradient of the classifier $\nabla_{\bm x}\log q_t(\bm y|\bm x_t)$ can be written to $\nabla_{\bm x}\log q_t(\bm x_t|\bm y)-\nabla_{\bm x}\log q_t(\bm x_t)$. By training a conditional noise prediction model $\bm\epsilon_\theta(\bm x_t, t, \bm y)$, the sampling process can be guided with no additional classifier:
\begin{equation}
    \bar{\bm\epsilon}_\theta(\bm x_t, t, \bm y)=\bm\epsilon_\theta(\bm x_t, t)-w\sigma_t\nabla_{\bm x}\log q_t(\bm y|\bm x_t)=\bm\epsilon_\theta(\bm x_t, t)+w(\bm\epsilon_\theta(\bm x_t, t, \bm y)-\bm\epsilon_\theta(\bm x_t, t))
\end{equation}
where $\bm\epsilon_\theta(\bm x_t, t)=\bm\epsilon_\theta(\bm x_t, t, \Phi)$ is approximated by the noise prediction model conditioned on a pre-specified label $\Phi$ standing for non-conditioning. Although CFG can generate trajectories specific to condition $\bm y$, it may cause the agent to reject higher likelihood trajectories in sequential environments, resulting in a performance drop \citep{pearce2023imitating}. Therefore, some methods~\cite{chi2023diffusion, pearce2023imitating, wang2023diffusion} set the guidance weight $w$ to 1, i.e., no guidance paradigm. 

\section{Related Works}

In recent years, DMs have demonstrated promising performance in various domains \citep{zhang2023controlnet, Ruiz2023dreambooth, Liu2023audioldm, kingma2021variational}, giving rise to several high-quality DM libraries, such as Diffusers \citep{von-platen-etal-2022-diffusers} and Stable Diffusion \citep{rombach2021highresolution}. These open-source libraries have significantly promoted research and applications in related fields. However, unfortunately, these libraries are designed for multimedia such as image, audio, and video generation, lacking adaptation for decision-making tasks. This is likely because DMs play diverse roles in decision-making, with various usage patterns and many unique mechanism incorporations, creating a gap in the multimedia generation paradigm. A library specially designed for decision-making is currently missing, and most research codebases are inherited from a few pioneering studies \citep{janner2022planning, wang2023diffusion, chi2023diffusion}. While effective, their algorithm-specific mechanisms and tightly coupled system architecture make it challenging for customized development.

\alg aims to provide an "easy-to-hack" starter kit for research needs, offering researchers more exploration possibilities. We draw from the experience of many open-source decision-making libraries. For example, we emulate stable-baselines3 \citep{stable-baselines3} to carefully reproduce results to provide practitioners with reliable baselines for method comparison. However, we inject more modular design to encourage users to freely design and modify. We also follow CORL \citep{tarasov2022corl} in designing clean and logically clear pipelines for readability, but, considering the complexity of DMs, abandon the one-file-from-scratch approach and opt for a one-file pipeline approach to offer rich examples of how to utilize \alg building blocks to implement decision-making algorithms. Additionally, we follow Ray \citep{liang2018rllib} in providing ample parameter selection interfaces within modules, making it easy for users unfamiliar with the internal implementation to customize effortlessly. In summary, \alg is not only the first open-sourced modularized DM library tailored for decision-making algorithms but also a new library that draws on the advanced experiences of many open-source decision-making libraries.

\section{Details of Experimental Setup}\label{append:env_details}
\subsection{Offline Reinforcement Learning Environments and Datasets}

\begin{figure}[h]
    \centering
    \includegraphics[width=0.9\textwidth]{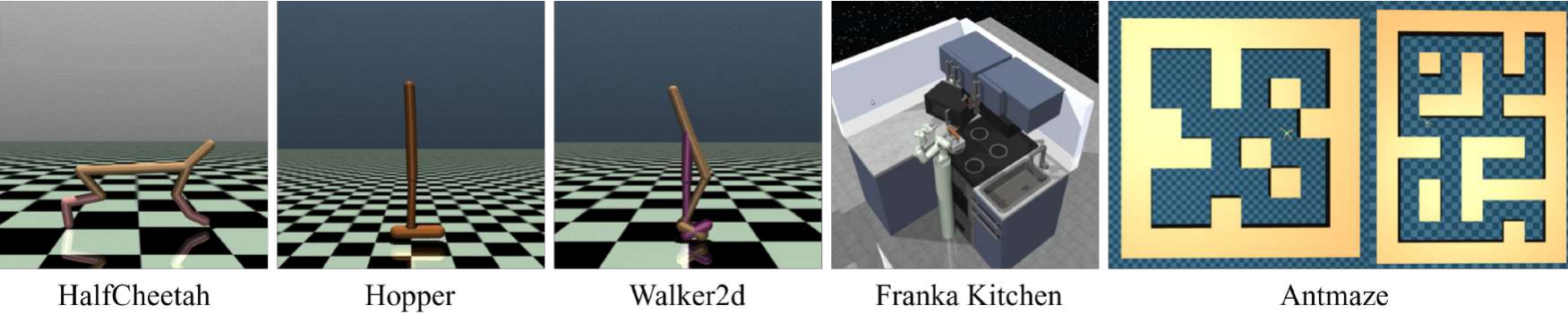}
    \caption{\small{\textbf{Visualization of Offline Reinforcement Learning Environments.}}}
    \label{fig:rl_benchmarks}
\end{figure}

We evaluate 7 diffusion-based RL algorithms implemented with CleanDiffuser on 15 offline RL tasks from 3 benchmarks, including locomotion, manipulation, and navigation. These tasks are widely recognized and extensively used in offline RL settings \citep{kumar2020cql, fujimoto2021td3bc, kostrikov2022iql, hansen2023idql, wang2023diffusion, kang2024efficient, janner2022planning, ajay2022conditional, dong2024diffuserlite, li2023hdmi, he2024diffcps}, enjoying significant acceptance within the research community. Visualization of these tasks is presented in \Cref{fig:rl_benchmarks}. These tasks come from the D4RL benchmark, in which the datasets are licensed under the Creative Commons Attribution 4.0 License (CC BY), and the code is licensed under the Apache 2.0 License.

\textbf{Gym-MuJoCo} \citep{brockman2016openai} consists of three popular offline RL locomotion tasks (HalfCheetah, Hopper, Walker2d), which require controlling three Mujoco robots to achieve maximum movement speed while minimizing energy consumption under stable conditions. D4RL \citep{fu2020d4rl} benchmark provides three different quality levels of offline datasets: ``medium'' containing demonstrations of medium-level performance; ``medium-replay'' containing all recordings in the replay buffer observed during training until the policy reaches ``medium'' performance; and ``medium-expert'' which combines ``medium'' and ``expert'' level performance equally.

\textbf{Franka Kitchen} \citep{gupta2019relay} requires controlling a realistic 9-DoF Franka robot arm to complete several household tasks in a kitchen environment. Algorithms are trained on ``partial'' and ``mixed'' datasets. The ``partial'' and ``mixed'' datasets consist of undirected data, where the robot performs subtasks that are not necessarily related to the goal configuration. In the ``partial'' dataset, a subset of the dataset is guaranteed to solve the task, meaning an imitation learning agent may learn by selectively choosing the right subsets of the data. The ``mixed'' dataset contains no trajectories that solve the task completely, and the RL agent must learn to assemble the relevant sub-trajectories. This dataset requires the highest degree of generalization in order to succeed.

\textbf{Antmaze} \citep{fu2020d4rl} requires controlling the 8-DoF ``Ant'' quadruped robot to complete maze navigation tasks. In the offline dataset, the robot only receives a reward upon reaching the goal, and the dataset contains many trajectory segments that do not lead to the endpoint, making it a difficult decision task with sparse rewards and a long horizon. The success rate of reaching the endpoint is used as the evaluation score, and common offline RL algorithms often struggle to achieve good performance.

\subsection{Offline Imitation Learning Environments and Datasets}

\begin{figure}[h]
\centering
\includegraphics[width=1.0\textwidth]{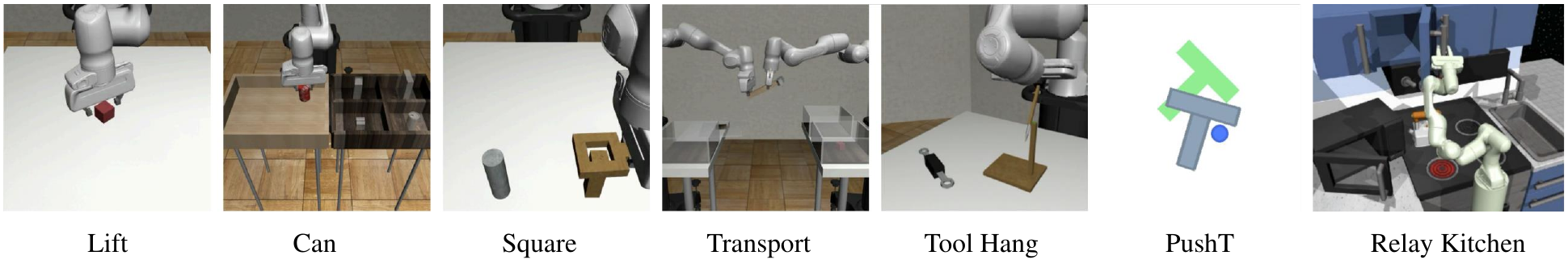}
\caption{\textbf{Visualization of Offline Imitation Learning Environments.}}
\label{fig:il_benchmark}
\end{figure}

We evaluate 2 diffusion-based IL algorithms implemented with CleanDiffuser on 22 imitation learning tasks from 4 benchmarks, with both state and image-based observation inputs. Among them, Relay Kitchen and Robomimic support both velocity and position control. Each algorithm is trained with its best-performing action space. We provide task summary in \Cref{tab:imitation_task_summary}, visualization in \Cref{fig:il_benchmark}, and more details below:

\textbf{PushT}~\cite{florence2022implicit} requires pushing a T-shaped block~(gray) to a fixed target~(red) with a circular end-effector. The task requires exploiting complex and contact-rich object dynamics to push the T block precisely, using point contacts. In this paper, we used three variants. ``PushT'' env has a five-dimensional state space, including the proprioception for end-effector location \texttt{(agent\_x,  agent\_y)} and the xy coordinates and angles of the blocks \texttt{(block\_x,  block\_y, block\_angle)}. ``PushT-keypoints'' env includes nine 2D key points obtained from the T-block's ground truth attitude and proprioception for end-effector location. ``Pusht-image'' env observes the end-effector location and the top view of the RGB image. This benchmark is licensed under the Apache-2.0 License.

\textbf{Relay Kitchen} is proposed in Relay Policy Learning~\cite{gupta2019relay}, commonly used to evaluate imitative learning ability. The environment consists of a 9 DoF position-controlled Franka robot interacting with a kitchen scene that includes an openable microwave, four turnable oven burners, an oven light switch, a freely movable kettle, two hinged cabinets,
and a sliding cabinet door. The ``relay'' dataset contains 566 human demonstrations, each completing four tasks in arbitrary order. The goal is to execute as many tasks as possible, regardless of order, showcasing both short-horizon and long-horizon multimodality. This benchmark is licensed under the Apache-2.0 License.

\textbf{Robomimic}~\cite{robomimic2021} requires controlling a robot arm to complete complex manipulation tasks from a few human demonstrations. Due to the non-Markovian nature of human demonstrations and the demonstration quality variance, learning from human datasets is significantly more challenging than learning from machine-generated datasets. Proficient-Human (PH) and Multi-Human (MH) datasets are collected by humans through remote teleoperation. The PH datasets consist of 200 demonstrations collected by a single, experienced teleoperator, while the MH datasets consist of 300 demonstrations collected by 6 teleoperators of varying proficiency, each of which provided 50 demonstrations. The benchmark consists of 5 PH tasks (Lift, Can, Square, Tool\_hang, Transport) and 4 MH tasks (Lift, Can, Square, Transport). Each task has both state and image-based observation inputs. This benchmark is licensed under the MIT License.

To the best of our knowledge, the datasets and benchmarks we have used do not contain personally identifiable information or offensive content in both previous works and our works.

\begin{table}[htb]
\caption{\small{\textbf{Imitation Learning Task Summary.} Obs Shape represents the low dimensional state space dimension; Image Shape represents the observation resolution of multi-view images (Camera views x W x H). PH: proficient-human demonstration, MH: multi-human
demonstration, Steps: max episode steps.}}
\scalebox{0.7}{
\begin{tabular}{@{}l|c|cc|cccc@{}}
\toprule
\multirow{2}{*}[-0.5ex]{Task} & 
  Low Dim Tasks &
  \multicolumn{2}{c|}{Image Tasks} &
  \multirow{2}{*}[-0.5ex]{Action Dim} & 
  \multirow{2}{*}[-0.5ex]{PH Demonstration} & 
  \multirow{2}{*}[-0.5ex]{MH Demonstration} & 
  \multirow{2}{*}[-0.5ex]{Max Steps} \\ 
\cmidrule(lr){2-2} \cmidrule(lr){3-4}  
 &
  Obs Shape &
  Obs Shape &
  Image Shape &
   &
   &
   &
   \\ \midrule
PushT          & 5   & N/A & N/A       & 2 & 200 & N/A & 300 \\
PushT-Keypoint & 20  & N/A & N/A       & 2 & 200 & N/A & 300 \\
PushT-Image    & N/A & 2   & 1x96x96   & 2 & 200 & N/A & 300 \\
Relay Kitchen  & 60  & N/A & N/A       & 9 & 656 & N/A & 280 \\
Lift           & 19  & 9   & 2x84x84   & 7 & 200 & 300 & 400 \\
Can            & 23  & 9   & 2x84x84   & 7 & 200 & 300 & 400 \\
Square         & 23  & 9   & 2x84x84   & 7 & 200 & 300 & 500 \\
Transport      & 59  & 18  & 4x84x84   & 7 & 200 & 300 & 700 \\
Tool\_hang     & 53  & 9   & 2x240x240 & 7 & 200 & N/A & 700 \\
\bottomrule
\end{tabular}
}
\label{tab:imitation_task_summary}
\end{table}

\section{Additional Experiments}\label{append:additional_exp}

\subsection{Impact of Model Size in RL Benchmarks}
\label{append:model_size}

\begin{table}[htb]
\caption{\small{\textbf{Impact of Model Size in RL Benchmarks.} Performance of DD and IDQL with varying model sizes. Results correspond to the mean and standard error over 150 episode seeds.}}
\centering
\label{tab:model_size}
\scalebox{0.8}{
\begin{tabular}{lccc|ccc}
\toprule
\textbf{Environment} & \multicolumn{3}{c|}{\textbf{DD}} & \multicolumn{3}{c}{\textbf{IDQL}} \\ \cmidrule(l){1-7} 
\textbf{Model Size} & 4M & 15M & 60M & 1.6M & 6M & 25M \\ \midrule
\textbf{HalfCheetah-m} & $45.3\pm0.3$ & $44.5\pm0.1$ & $\bm{47.1\pm0.1}$ & $51.5\pm0.1$ & $51.5\pm0.1$ & $\bm{51.7\pm0.1}$ \\
\textbf{Kitchen-m} & $56.5\pm5.8$ & $\bm{80.5\pm4.1}$ & $27.7\pm2.1$ & $66.5\pm4.1$ & $\bm{69.2\pm1.0}$ & $67.5\pm1.8$ \\
\textbf{Antmaze} & \multirow{2}{*}{$8.0\pm4.3$} & \multirow{2}{*}{$\bm{26.0\pm5.9}$} & \multirow{2}{*}{$22.7\pm6.6$} & \multirow{2}{*}{$48.7\pm4.7$} & \multirow{2}{*}{$52.0\pm5.7$} & \multirow{2}{*}{$\bm{54.0\pm4.3}$} \\ 
\textbf{(mp for DD, lp for IDQL)} & & & & & & \\
\bottomrule
\end{tabular}}
\end{table}

There is a significant disparity in network model sizes used by diffusion-based decision-making algorithms. For instance, the official implementation of DD utilizes around 60M parameters \citep{ajay2022conditional}, while Diffuser uses 4M \citep{janner2022planning}, and IDQL \citep{hansen2023idql} has approximately only 1.6M parameters. These works have limited discussion on the impact of model size. Therefore, we aim to explore the approximate scale of parameter sizes required for diffusion-based decision-making algorithms to function effectively. In this experiment, we test DD and IDQL at three different model sizes, starting from the default parameter size used in the main experiments and gradually increasing the parameter size by four times. The performance of the algorithms is evaluated on three tasks including locomotion, manipulation, and navigation. Results are presented in \Cref{tab:model_size}. We find that, apart from the performance of DD on Kitchen-m and Antmaze-mp, increasing the model size does not lead to significant performance gains in other cases. However, even with the performance gains brought by model size,  DD can not entirely catch up with the performance of IDQL, indicating that the dominant effect on performance is still primarily driven by the algorithm rather than the model size.

\subsection{Impact of Diffusion Backbones and Sampling Steps (Full Results)}\label{append:full_ss_solver}

\begin{figure}[htb]
\centering
\includegraphics[width=1.0\textwidth]{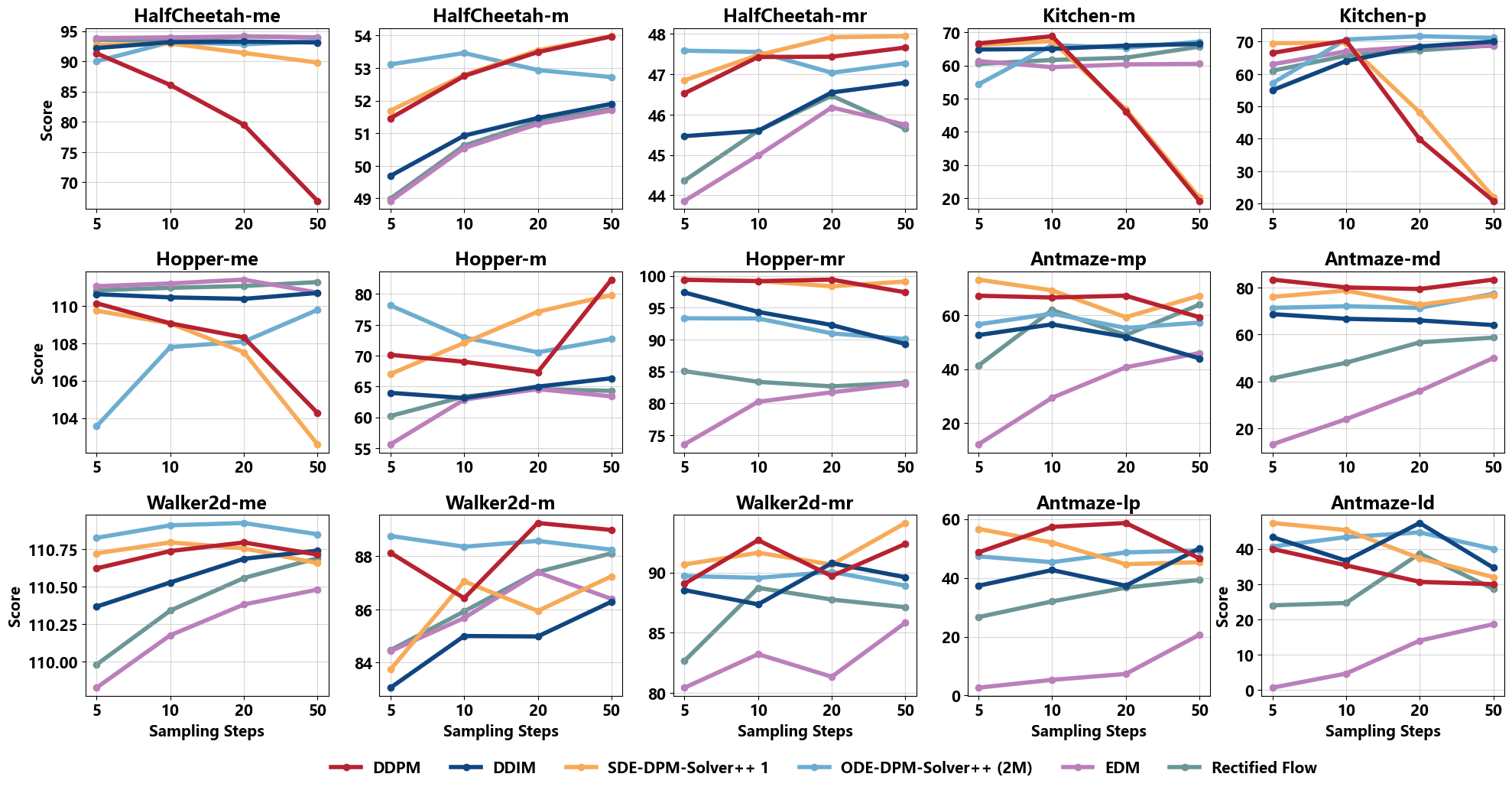}
\caption{\small{\textbf{Full D4RL Results of IDQL.} Performance of IDQL with various diffusion backbones and varying sampling steps. Results correspond to the mean over 150 episode seeds.}}
\label{fig:idql_ss_solver}
\end{figure}

\begin{figure}[htb]
\centering
\includegraphics[width=1.0\textwidth]{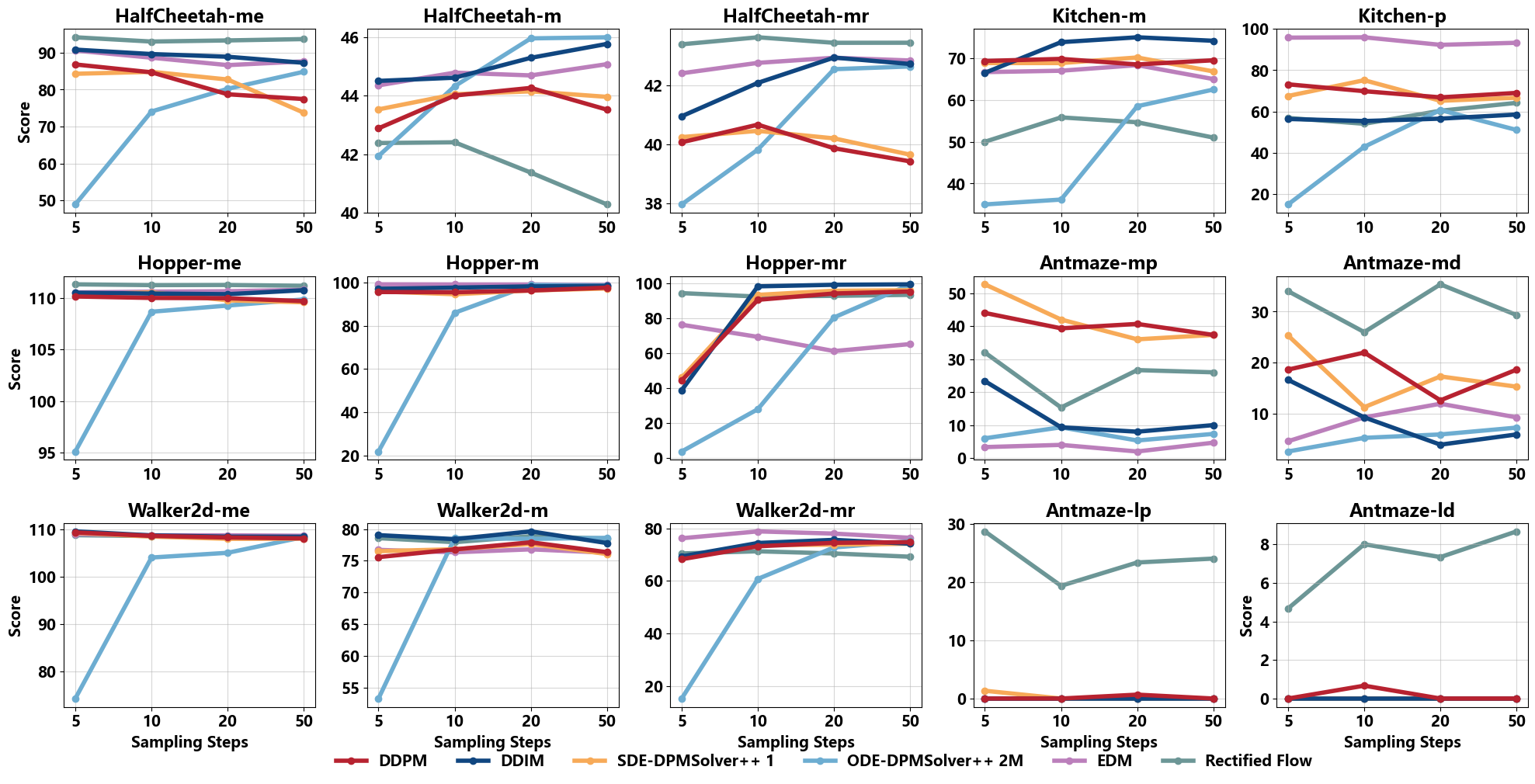}
\caption{\small{\textbf{Full D4RL Results of DD.} Performance of DD with various diffusion backbones and varying sampling steps. Results correspond to the mean over 150 episode seeds.}}
\label{fig:dd_ss_solver}
\end{figure}

Due to space limitations in the main text, we present the full results of IDQL and DD on D4RL in \Cref{fig:idql_ss_solver} and \Cref{fig:dd_ss_solver}. The algorithms are trained for $1\times10^6$ gradient steps, and the sampling steps for DD are set to 5, with other hyperparameters consistent with default settings. This experiment selects DDPM, DDIM, SDE-DPM-Solver++ 1, ODE-DPM-Solver++ (2M), EDM, and Rectified Flow as the diffusion/solver backbones. We select DDPM and DDIM because they are the first-order discretization of diffusion reverse SDE/ODE, respectively \citep{song2021scorebased, song2021ddim}. We do not choose DPM-Solver because its first-order solver is equivalent to DDIM \citep{lu2022dpmsolver}, and higher-order solvers may cause instability under guidance \citep{lu2023dpmsolverpp}. For DPM-Solver++, we select a first-order SDE solver, SDE-DPM-Solver++ 1, and a second-order ODE solver, ODE-DPM-Solver++ (2M). Since higher-order solvers can lead to instability, they are therefore not chosen. We select EDM and Rectified Flow because they have achieved excellent results in image generation but have not been widely used in the decision-making domain, to the best of our knowledge. 
Thanks to \alg's support for various solvers and varying sampling steps, the results for DDPM, DDIM, SDE-DPM-Solver++ 1, and ODE-DPM-Solver++ (2M) only require training one single model. Additionally, using different sampling steps does not require additional training. These features provide a great convenience for conducting ablation experiments. We believe these features of \alg can also benefit future research efforts.

\subsection{Additional Analyses of DMs in IL Benchmarks}
\label{app:additional_exp}

\begin{wraptable}{r}{0.5\textwidth}
\vspace{-18pt}
\caption{\small{\textbf{The Model Size and Inference Time of DiffusionPolicy and DiffusionBC in Low-Dim Lift-ph.} DiffusionPolicy uses 50 sampling steps across the experiments, and DiffusionBC incorporates 8 additional Diffusion-X sampling steps.}}
\small
\centering
\label{tab:model_size_table}
\resizebox{0.48\textwidth}{!}{%
\begin{tabular}{@{}lll@{}}
\toprule
Algorithm                                                                     & Model Size (M) & Inference Time (s) \\ \midrule
\begin{tabular}[c]{@{}l@{}}DiffusionPolicy\\ w/ \texttt{Chi\_UNet1d}\end{tabular}      & 68.91          & 0.405                   \\
\begin{tabular}[c]{@{}l@{}}DiffusionPolicy\\ w/ \texttt{Chi\_TFM}\end{tabular} & 9.50               & 0.343                   \\
\begin{tabular}[c]{@{}l@{}}DiffusionPolicy\\ w/ \texttt{DiT1d}\end{tabular}            & 16.59           & 0.194                   \\ \midrule
\begin{tabular}[c]{@{}l@{}}DiffusionBC\\ w/ \texttt{DiT1d}\end{tabular}                & 16.59           & 0.217                   \\
\begin{tabular}[c]{@{}l@{}}DiffusionBC\\ w/ \texttt{Pearce\_MLP}\end{tabular}          & 0.83           & 0.062                   \\ \midrule
ACT                                                                           & 7.83           & 0.006                   \\ \bottomrule
\end{tabular}
}
\vspace{-5pt}
\end{wraptable}

Using the low-dim lift-ph task with 50 sample steps in Robomimic as a reference, we present the number of parameters and inference time for each variant of DiffusionPolicy, DiffusionBC, and ACT in \cref{tab:model_size_table}. Although \texttt{Chi\_UNet1d} exhibits the best performance in many IL tasks, it has the largest model size and the slowest inference speed. Larger model size results in higher training costs, and in many real-world applications that require real-time inference, we need to make trade-offs between inference speed and performance. Compared to the transformer-based ACT algorithm, all structures of the diffusion policy exhibit slower sampling speeds because the denoising process requires multiple forwards for neural networks. This is also an important challenge that limits the application of DMs for decision-making. We also note that DiffusionBC is slower than DiffusionPolicy when using the same network architecture and model size, as DiffusionBC performs 8 additional steps of Diffusion-X sampling to mitigate OOD issues. Although the best-performing \texttt{Chi\_UNet1d} model uses a considerable model size, simply increasing the Transformer-based DMs like \texttt{DiT1d} can sometimes harm performance. We discuss this in detail in \cref{append:model_size}, which is also consistent with the experimental observations of the \cite{chi2023diffusion}. Finding the optimal model size in applications remains an open research question.

\section{Experimental Details}

\subsection{Computing Resources}\label{append:computing_resources}

RL experiments are conducted on a server equipped with 2 Intel(R) Xeon(R) Gold 6326 CPUs @ 2.90GHz and 8 NVIDIA GeForce RTX3090 GPUs, and a server equipped with 2 Intel(R) Xeon(R) Gold 6326 CPUs @ 2.90GHz and 8 NVIDIA GeForce RTX2080Ti GPUs.
IL experiments are conducted on a server equipped with 2 Intel(R) Xeon(R) Gold 6338 CPUs @ 2.00GHz and 8 NVIDIA A800 GPUs, and a server equipped with 2 Intel(R) Xeon(R) Gold 6338 CPUs @ 2.00GHz and 4 NVIDIA GeForce RTX3090 GPUs.

\subsection{Evaluation Metircs}\label{append:metrics}

In the D4RL benchmark, the scores
are normalized to the range between 0 and 100 with expert-normalized scores $=100 \times \frac{\text { score } \times \text { random\_score }}{\text { expert\_score-random\_score }}$~\cite{fu2020d4rl}. As for IL benchmarks, we report target area coverage as scores in the PushT benchmark and success rate in the Robomimic benchmark. In the Relay Kitchen environment, since the vast majority of human demonstrations can only complete 4 subtasks, we denote the success rate of completing the $i$-th subtask as $p_i$ and report the average success rate as $\text{score} = {(p_1 + p_2 + p_3 + p_4)}/{4}$.

\subsection{Algorithm Hyperparameters}\label{append:hyperparams}

Unless stated otherwise, we utilize default hyperparameters from the official implementations for most algorithms and datasets. \textbf{Configuration files and hyperparameters for each algorithm and environment are available in YAML format on our GitHub repository for reproducibility.}

Key hyperparameters for each offline RL algorithm are presented in \Cref{tab:rl_hyperparams}, and each offline IL algorithm in \Cref{tab:il_hyperparams}. We also reproduce the Transformer-based ACT~\cite{zhao2023learning} algorithm based on the official implementation, the key hyperparameters are in Table 50000.



\begin{table}[h]
\caption{\small{\textbf{Hyperparameters for Diffusion Planners, Diffusion Policies and Diffusion Data Synthesizer for RL.}}}
\small
\centering
\label{tab:rl_hyperparams}
\begin{small}
\scalebox{0.77}{
\begin{tabular}{l|ccc|ccc|c}
\toprule
\textbf{Hyperparameter} & \textbf{Diffuser} & \textbf{DD} & \textbf{AdaptDiffuser} & \textbf{DQL} & \textbf{EDP} & \textbf{IDQL} & \textbf{SynthER}\\ 
\midrule
Architecture & Janner\_UNet & DiT  & Janner\_UNet & DQL\_MLP & DQL\_MLP & LNResnet & LNResnet \\
Diffusion Model & DDPM & DDIM & DDPM & DDPM & DPM-Solver++~(2M) & DDPM & DDIM\\
Sampling Steps & 20 & 20 & 20 & 5 & 15 & 5 &  128 \\
Horizon & 64 (Antmaze) & 64 (Antmaze) & 64 (Antmaze) & 1 & 1 & 1 & 1\\
& 32 (Otherwise) & 32 (Otherwise) & 32 (Otherwise) & & & & \\
Temperature & 0.5 & 0.5 & 0.5 & 0.5 & 0.5 & 0.5 & 1.0\\
Gradient Steps & 1e6 & 1e6 & 1e6 & 2e6 & 2e6 & 2e6 & 1e5\\
Batch Size & 64 & 64 & 64 & 256 & 256 & 256 & 256\\
Learning Rate & 3e-4 & 3e-4 & 3e-4 & 3e-4 & 3e-4 & 3e-4 & 3e-4 \\
N candidates & 64 & 1 & 64 &  50 & 50 & 256 & N/A \\
\bottomrule
\end{tabular}}
\end{small}
\end{table}

\begin{table}[htb]
\caption{\small{\textbf{Hyperparameters for DiffusionPolicy and DiffusionBC in Low-Dim and Image Tasks.}}}
\small
\centering
\label{tab:il_hyperparams}
\scalebox{0.92}{
\begin{tabular}{l|ccc|cc}
\toprule
\textbf{Hyperparameters} & \textbf{} & \textbf{DiffusionPolicy} & \textbf{} & \multicolumn{2}{c}{\textbf{DiffusionBC}} \\
\midrule
Architecture             & \texttt{Chi\_UNet1d} & \texttt{Chi\_Transformer}         & \texttt{DiT1d}       & \texttt{Pearce\_MLP}             & \texttt{DiT}            \\
Diffusion Model    & DDPM & DDPM & DDPM  & DDPM & DDPM \\
Sampling Steps     & 5~(PushT)   & 5~(PushT)   & 5~(PushT)   & 50   & 50   \\
                   & 50~(Otherwise)   & 50~(Otherwise)   & 50~(Otherwise)   &    &   \\
Horizon            & 16   & 10   & 10   & 2    & 2    \\
Obs Steps          & 2    & 2    & 2    & 2    & 2    \\
Action Steps       & 8    & 8    & 8    & 1    & 1    \\
Gradient Steps     & 1e6  & 1e6  & 1e6  & 1e6  & 1e6  \\
Batch Size         & 256~(Low dim)  & 256~(Low dim)  & 256~(Low dim)  & 512~(Low dim)  & 512~(Low dim)  \\
                   & 64~(Image)  & 64~(Image)  & 64~(Image)  & 64~(Image)  & 64~(Image)  \\
Temperature        & 1.0  & 1.0  & 1.0  & 1.0  & 1.0  \\
Learning Rate      & 1e-4 & 1e-4 & 1e-4 & 1e-3 & 5e-4 \\
Extra Sample Steps & N/A  & N/A  & N/A  & 8    & 8    \\ 
Control Mode       & Pos  & Pos  & Pos  & Vel  & Vel    \\ 
\bottomrule
\end{tabular}
}
\end{table}

\begin{table}[h!]
\caption{\small{\textbf{Hyperparameters for ACT in Low-Dim and Image Tasks.}}}
\small
\centering
\begin{tabular}{ll}
\toprule
\textbf{Hyperparameters} & \textbf{Value} \\
\midrule
Learning Rate & 1e-5 \\
Batch Size & 256 (Low dim) / 64 (Image) \\
\# Encoder Layers & 4 \\
\# Decoder Layers & 7 \\
Feedforward Dimension & 256 \\
Hidden Dimension & 256 \\
\# Heads & 8 \\
Chunk size & 16 \\
Beta & 10 \\
Gradient Steps & 1e6 \\
Control Mode & Vel (Kitchen) / Pos (Otherwise) \\
\bottomrule
\end{tabular}
\end{table}

\section{Implemented Diffusion Models}\label{append:implemented_dms}
\subsection{DDPM/DDIM/DPM-Solver/DPM-Solver++}

\textbf{Applying Solvers with One Score Function.}
Due to the generation processes of DDPM \citep{ho2020ddpm}, DDIM \citep{song2021ddim}, DPM-Solver \citep{lu2022dpmsolver}, and DPM-Solver++ \citep{lu2023dpmsolverpp} can all be expressed using the same diffusion SDE/ODE \citep{song2021scorebased}, utilizing the same noise schedule, training just one noise predictor model enables the use of these four solvers for sampling. Recall that the diffusion ODE with noise prediction model is:
\begin{equation}
    \frac{\dd x_t}{\dd t}=f(t)\bm x_t+\frac{g^2(t)}{2\sigma_t}\bm\epsilon_\theta(\bm x_t, t).
\end{equation}
Substituting $f(t)=\frac{\dd\log\alpha_t}{\dd t}, g^2(t)=\frac{\dd\sigma^2_t}{\dd t}-2\sigma^2_t\frac{\dd\log\alpha_t}{\dd t}$, and conducting first-order discretization result in a recursive formula:
\begin{align}
    \bm x_{t}-\bm x_s &= \frac{\alpha_{t}-\alpha_s}{\alpha_s}\bm x_s+\frac{1}{2\sigma_s}\left[2\sigma_s(\sigma_{t}-\sigma_s)-2\frac{\sigma^2_s}{\alpha_s}(\alpha_{t}-\alpha_s)\right]\bm\epsilon_\theta(\bm x_s) \\
    \bm x_t &=\frac{\alpha_t}{\alpha_s}\bm x_s - \alpha_t\left(\frac{\sigma_s}{\alpha_s}-\frac{\sigma_t}{\alpha_t}\right)\bm\epsilon_\theta(\bm x_s, s) \\
    \bm x_t &=\alpha_t\left(\frac{\bm x_t-\sigma_t\bm\epsilon_\theta(\bm x_s, s)}{\alpha_s}\right)+\sqrt{\sigma_s^2}\epsilon_\theta(\bm x_s, s),\label{eq:ddim}
\end{align}
where $t$ and $s$ are the next and current sampling steps. \Cref{eq:ddim} is DDIM update \citep{song2021ddim}. By introduce $\beta_s = (\sigma_t/\sigma_s)\sqrt{1-\alpha^2_s/\alpha^2_t}$, the generative process of DDPM is:
\begin{equation}
    \bm x_t =\alpha_t\left(\frac{\bm x_t-\sigma_t\bm\epsilon_\theta(\bm x_s, s)}{\alpha_s}\right)+\sqrt{\sigma_s^2-\beta^2_s}\epsilon_\theta(\bm x_s, s)+\beta_s\bm\epsilon_s,
\end{equation}
where $\bm\epsilon_s\sim\mathcal N(\bm 0,\bm I)$ is standard Gaussian noise independent of $\bm x_s$.
DPM-Solver leverages the semi-linearity of the diffusion ODE and formulates the exact solution by the ``variation of constants'' formula:
\begin{align}
    \bm x_t&=e^{\int_s^tf(\tau)\dd\tau}\bm x_s +\int_s^t\left(e^{\int_\tau^tf(r)\dd r}\frac{g^2(\tau)}{2\sigma_\tau}\bm\epsilon_\theta(\bm x_\tau,\tau)\right)\dd\tau \\
    \bm x_t &= \frac{\alpha_t}{\alpha_s}\bm x_s-\alpha_t\int_s^t\frac{\dd\lambda_\tau}{\dd\tau}\frac{\sigma_\tau}{\alpha_\tau}\bm\epsilon_\theta(\bm x_\tau, \tau)\dd\tau,
\end{align}
where $\lambda_t:=\log(\alpha_t/\sigma_t)$ is the log-\textit{signal-to-noise-ratio} (log-SNR). This formulation eliminates the approximation error of the linear term since it is exactly computed, and the non-linear term can be approximated using its Talor expansion:
\begin{equation}
    \bm x_t = \frac{\alpha_t}{\alpha_s}\bm x_s-\alpha_t\sum_{n=0}^{k-1}\bm\epsilon^{(n)}_\theta(\bm x, s)\int_{\lambda_{s}}^{\lambda_t}e^{-\lambda}\frac{(\lambda-\lambda_s)^n}{n!}\dd\lambda+\mathcal{O}((\lambda_t-\lambda_s)^{k+1}). \label{eq:dpmsolver}
\end{equation}
In \alg, we have implemented only \textit{DPM-Solver-1}, corresponding to the k=1 scenario in \Cref{eq:dpmsolver}, as guided sampling tends to make high-order solvers unstable \citep{lu2023dpmsolverpp}, leading to poor performance in decision-making tasks. DPM-Solver++ alleviates this instability issue by using a data prediction model $\bm x_\theta(\bm x_t, t)$ instead of the noise prediction model $\bm \epsilon_\theta(\bm x_t, t)$, transforming the generative process into:
\begin{equation}
    \bm x_t = \frac{\sigma_t}{\sigma_s}\bm x_s+\sigma_t\sum_{n=0}^{k-1}\bm x^{(n)}_\theta(\bm x, s)\int_{\lambda_{s}}^{\lambda_t}e^{\lambda}\frac{(\lambda-\lambda_s)^n}{n!}\dd\lambda+\mathcal{O}((\lambda_t-\lambda_s)^{k+1}), \label{eq:dpmsolverpp}
\end{equation}
where $\bm x_\theta(\bm x_t, t)$ is trained to predict the original data $\bm x_0$ from the perturbed data $\bm x_s$. In \alg, we have implemented DPM-Solver++ for $k\le2$, as it already yields satisfactory results at $k=2$, while higher-order solvers may still lead to instability.

Although the data prediction model can mitigate the instability issue caused by guided sampling and easily clip data to address the ``train-test mismatch'' problem \citep{lu2023dpmsolverpp}, there is still no definitive evidence in practice to determine the superiority of either the data prediction model or the noise prediction model. In \alg, we provide users with the option to choose between these two prediction models and use the approximation $\bm x_t\approx \alpha_t\bm x_\theta(\bm x_t, t)+\sigma\bm\epsilon_\theta(\bm x_t, t)$ to seamlessly switch between the two formulations to cater to the requirements of different solvers.

\textbf{Noise Schedules.} \alg provides two popular noise schedules by default: \textit{Linear Noise Schedule} \citep{ho2020ddpm} and \textit{Cosine Noise Schedule} \citep{nichol2021improved}. The former defines:
\begin{equation}
    \alpha_t = \exp\left(-\frac{(\beta_1-\beta_0)}{4}t^2-\frac{\beta_0}{2}t\right),
\end{equation}
where $\beta_0=0.1$, $\beta_1=20$ and $\sigma_t=\sqrt{1-\alpha_t^2}$. The diffusion SDE/ODE is solved between $[\epsilon, T]$, where $\epsilon=0.001$ and $T=1$ for numerical stability. The later schedule defines:
\begin{equation}
    \alpha_t=\frac{\cos\left(\frac{\pi}{2}\cdot\frac{t+s}{1+s}\right)}{\cos\left(\frac{\pi}{2}\cdot\frac{s}{1+s}\right)}
\end{equation}
where $s=0.008$ and $\sigma_t=\sqrt{1-\alpha_t^2}$. The diffusion SDE/ODE is solved between $[\epsilon, T]$, where $\epsilon=0.001$ and $T=0.9946$ for numerical stability. Beyond the two schedules, \alg allows users to fully customize new noise schedules according to the specified format to explore algorithm performance.


\subsection{EDM}

EDM \citep{karras2022edm} rewrites the diffusion forward process in \Cref{eq:forward} as:
\begin{equation}
    \bm x_t=s_t(\bm x_0 + \sigma_t \bm\epsilon_t),
\end{equation}
which can be interpreted as adding noise to a scaled version of the original data. By setting the scale $s_t\equiv1$ to a constant, EDM obtains the following reverse process:
\begin{equation}
    \frac{\dd\bm x_t}{\dd t}=-\dot\sigma_t\sigma_t\nabla_{\bm x}\log p(\bm x; \sigma_t)\dd t, 
\end{equation}
where $p(\bm x; \sigma_t)=p_t(\bm x)$. A data prediction model $D_\theta(\bm x; \sigma)$ is trained to approximate $\bm x+\sigma^2\nabla_{\bm x}\log p(\bm x;\sigma)$ and results in a practical generative process:
\begin{equation}
    \bm x_t = \bm x_s + (t-s)\cdot \left(\frac{\dot\sigma_s}{\sigma_s}\bm x_s-\frac{\dot\sigma_s}{\sigma_s}D_\theta(\bm x_s; \sigma_s)\right).
\end{equation}
One feature of EDM is that it applies preconditioning to $D_\theta$:
\begin{equation}
    D_\theta(\bm x; \sigma)=c_{\text{skip}}(\sigma)\bm x+c_{\text{out}}(\sigma)F_\theta(c_{\text{in}}(\sigma)\bm x; c_{\text{noise}}(\sigma)).
\end{equation}
where $F_\theta$ is the neural network to be trained, $c_{\text{skip}}$ modulates the skip connection, $c_{\text{in}}$ and $c_{\text{out}}$ scale the input and output magnitudes, and $c_{\text{noise}}$ maps noise level $\sigma$ into a conditioning input for $F_\theta$. $F_\theta$ is trained by minimizing the noising score matching loss:
\begin{equation}
    \mathcal L(\theta;\sigma)=\mathbb E_{\bm y\sim p_{\text{data}},\bm n\sim\mathcal N(\bm 0,\sigma^2\bm I)}\left[\lambda(\sigma)\Vert D_\theta(\bm y+\bm n;\sigma)-\bm y\Vert^2_2\right],
\end{equation}
where $\lambda(\sigma)$ is the loss weight. These coefficients are optimized to achieve the following objectives: (1) inputs of $F_\theta$ have unit variance, (2) training target of $F_\theta$ have unit variance, (3) $c_{\text{skip}}$ can minimize $c_{\text{out}}$ so that the errors of $F_\theta$ are amplified as little as possible, and (4) the loss of $F_\theta$ has a uniform weight across noise levels. The optimization results give the following design choices: $c_{\text{skip}}=\sigma^2_{\text{data}}/(\sigma^2+\sigma^2_{\text{data}})$, $c_{\text{out}}=\sigma\cdot\sigma_{\text{data}}/\sqrt{\sigma^2_{\text{data}}+\sigma^2}$, $c_{\text{in}}=1/\sqrt{\sigma^2_{\text{data}}+\sigma^2}$, $c_{\text{noise}}=\log(\sigma)/4$, and $\lambda(\sigma)=(\sigma^2_{\text{data}}+\sigma^2)/(\sigma_{\text{data}}\cdot\sigma)^2$. 

\textbf{Noise Schedule.} \alg provides only one default noise schedule, which is specially designed for EDM:
\begin{equation}
    \sigma_t = t, ~t\in\left[\sigma_{\text{min}}, \sigma_{\text{max}}\right], \label{eq:rf}
\end{equation}
where $\sigma_{\text{min}}=0.002$ and $\sigma_{\text{max}}=80$.

\subsection{Rectified Flow}
Rectified flow \citep{liu2023rectifiedflow} is an ODE on time $t\in[0, 1]$:
\begin{equation}
    \frac{\dd\bm x_t}{\dd t}=\bm v_\theta(\bm x_t, t),
\end{equation}
where the drift force $\bm v_\theta$ is trained to drive the flow to follow the direction $(\bm x_0-\bm x_1)$ of the linear path pointing from $\bm x_1$ to $\bm x_0$ as much as possible, by solving a simple least squares regression problem:
\begin{equation}
    \mathcal{L}(\theta)=\mathbb E_{\bm x_0\sim p_0, \bm x_1\sim p_1, t\sim\text{Uniform}(0,1)}\left[\left\Vert (\bm x_0-\bm x_1)-\bm v_\theta(\bm x_t, t)\right\Vert^2_2\right],
\end{equation}
where $\bm x_t=t\bm x_1+(1-t)\bm x_0$. It achieves the mutual transformation of samples from two distributions $p_0$ and $p_1$, by solving \Cref{eq:rf} forward or backward. Rectified flow possesses many favorable properties that allow it to continuously learn from its own sampled data to straighten the ODE flow, and this procedure is called \textit{reflow}. The straighter the ODE flow, the fewer sampling steps are needed to achieve good generation quality. In an ideal scenario, if the flow becomes completely straight, then we have:
\begin{equation}
    \bm x_t=t\bm x_1+(1-t)\bm x_0=\bm x_1+(1-t)\bm v(\bm x_1, 1),~\forall t\in[0,1],
\end{equation}
which enables one-step sampling.
The Rectified Flow implemented in \alg has full functionality to transform samples from any two arbitrary probability distributions. By default, it follows the settings in diffusion models, where $p_0$ is the dataset distribution and $p_1$ is the standard Gaussian distribution.

\section{Implemented Algorithms}\label{append:algos}
\subsection{Diffusion Planners}

\textbf{Diffuser.}~\cite{janner2022planning} Diffuser is the first diffusion planning algorithm, and its paradigm has been widely adopted in subsequent diffusion planning algorithms. Diffuser generates state-action pair trajectories $\bm x=[x^\tau,\cdots,x^{\tau+H-1}]$ from:
\begin{equation}
    p(\bm x|\mathcal O^{\tau:\mathcal T})\propto p(\bm x)p(\mathcal O^{\tau:\mathcal T}|\bm x)=p(\bm x)\prod_{t=\tau}^{\mathcal T}\exp(r(s^t,a^t)),
\end{equation}
where $\mathcal O^{t_1:t_2}$ is a binary random variable denoting the optimality of a trajectory from $t_1$ to $t_2$, and $\mathcal T$ is the episode terminal time step of the trajectory \footnote{In previous works, authors typically consider only the trajectory cumulative reward as the generative condition, i.e. using $\mathcal O^{\tau:\tau+H-1}$, which overlooks future optimality. Their code implementations actually use the episodic cumulative reward, i.e. $\mathcal O^{\tau:\mathcal T}$. Therefore, we adopt this episodic cumulative reward expression.}. Therefore, it is natural to define the classifier in CG as a reward function on perturbed trajectories:
\begin{equation}
    \nabla_{\bm x}\log p_t(\bm x_t|\mathcal O^{\tau:\mathcal T})=\nabla_{\bm x}\log p_t(\bm x_t)+\sum_{k=\tau}^{\mathcal T}\nabla_{s_t^k,a_t^k}r(s_t^k,a_t^k)=\nabla_{\bm x}\log p_t(\bm x_t)+\nabla_{\bm x}\mathcal J_\phi(\bm x_t,t),
\end{equation}
where $\mathcal J_\phi(\bm x_t, t)$ is a neural network trained to predict the episodic cumulative reward $\sum_{k=\tau}^{\mathcal T}r(s_t^k,a_t^k)$ of the perturbed trajectory $\bm x_t$. At each inference step, given the current state $s^k$, Diffuser sets and freezes the first state of the trajectory as $s^k$ and performs guided sampling in an inpainting manner to generate a set of trajectories $\{\bm x_0\}$. Subsequently, it identifies the optimal trajectory $\bm x_0^*=\arg\max_{\bm x_0} \mathcal J_\phi(\bm x_0, 0)$ that maximizes the episodic cumulative reward, and extracts the first action $a^k$ in $\bm x_0^*$ to execute.

\textbf{Decision Diffuser.}~\cite{ajay2022conditional} Decision Diffuser (DD) introduces another prominent framework that utilizes a state-only trajectory formulation and implements CFG by discarding the optimality variable $\mathcal O^{\tau:\mathcal T}$ in favor of directly employing normalized episodic cumulative reward $y=\sum_{t=\tau}^{\mathcal T}r(\bm x)$ as the condition. As no additional reward predictor can be used for trajectory selection, DD generates only a single trajectory at each inference step and employs an trained inverse dynamic model $\mathcal I_\phi$ to predict the action to be executed $a^t=\mathcal I_\phi(s^t,s^{t+1})$.

\textbf{AdaptDiffuser.}~\cite{liang2023adaptdiffuser} Observing that the insufficient diversity of offline RL training data may limit the sample quality of DMs, AdaptDiffuser, an extension of Diffuser, proposes to utilize self-generated diverse synthetic expert data to fine-tune itself. The pipeline of AdaptDiffuser involves initially training a Diffuser as usual, then generating a large amount of synthetic expert data and using a discriminator to filter out high-quality data. Finally, fine-tuning is done on this dataset. This self-evolving process can be repeated multiple times to optimize the model, and different directions of model self-evolution can be controlled by designing different discriminators. The inference method of AdaptDiffuser is consistent with Diffuser, and its performance for seen tasks has been enhanced while also being able to adapt to unseen tasks.

\subsection{Diffusion Polices}
\textbf{Diffusion Q-Learning.}~\cite{wang2023diffusion} Diffusion Q-learning (DQL) leverages the capability of DMs to model complex distributions, directly applying DDPM as the policy $\pi_\theta(\bm a_0|\bm s)$ in the RL actor-critic framework. Sampling from the policy is therefore equivalent to the denoising process of the diffusion model. The Bellman operator can be used to train the Q-value function of the diffusion policy:
\begin{equation}
    \mathcal{L}(\phi)=\mathbb E_{(\bm s^k,\bm a^k,r,\bm s^{k+1})\sim\mathcal D, \bm a^{k+1}_0\sim\pi_{\theta'}}\left[\left\Vert(r+\gamma\min_{i=1,2}Q_{\phi_i'}(\bm s^{k+1}, \bm a^{k+1}_0))-Q_{\phi_i}(\bm s^k,\bm a^k)\right\Vert^2_2\right],
\end{equation}
where $\phi_1$ and $\phi_2$ represent the parameters of the double Q-learning trick, $\phi'$ and $\theta'$ represent the target networks. For policy optimization, DQL employs the most basic form of Offline RL optimization, which involves training the policy to maximize the Q-value while imitating behavior policies, using a weighting factor $\alpha$ to balance the influence of both aspects:
\begin{equation}
    \mathcal{L}(\theta)=\mathcal{L}_{\text{score}}(\theta)-\alpha\cdot\mathbb E_{\bm s\sim\mathcal D, \bm a_0\sim\pi_\theta}\left[Q_\phi(\bm s, \bm a_0)\right],
\end{equation}
where $\mathcal{L}_{\text{score}}(\theta)$ is the score matching loss used for diffusion model training. As the scale of the Q-value function varies in different offline datasets, to normalize it, DQL sets $\alpha=\frac{\eta}{\mathbb E_{(\bm s,\bm a)\sim\mathcal D}[|Q_\phi(\bm s,\bm a)|]}$ and tunes $\eta$ for loss term balance. The $Q_\phi$ in the denominator is only for normalization and not differentiated over.

\textbf{Efficient Diffusion Policy.}~\cite{kang2024efficient} Efficient Diffusion Policy (EDP) aims to address the significant computational overhead caused by iterative sampling and gradient computation during the training of the DQL. Compared to DQL, EDP proposes using DPM-Solver instead of DDPM to reduce the number of sampling steps. Then, EDP introduces an \textit{action approximation} technique, where during policy optimization, one-step denoising is performed on the perturbed action $\bm a_t$ to approximate $\bm a_0$. For the process using a data prediction model $\bm x_\theta$ and a noise prediction model $\bm\epsilon_\theta$ separately, the following two equations can express the technique:
\begin{align}
    \bm a_0 &\approx \bm x_\theta(\bm a_t, t) \\
    \bm a_0 &\approx \frac{\bm a_t - \sigma_t\bm\epsilon_\theta(\bm a_t, t)}{\alpha_t}.
\end{align}
EDP reduces the sampling steps to 15 (even though DQL has only 5 sampling steps) and performs only one-step denoising during policy optimization, significantly speeding up the model training process and achieving performance close to that of DQL.

\textbf{Implicit Diffusion Q-Learning.}~\cite{hansen2023idql} Implicit Diffusion Q-Learning (IDQL) models the policy from the perspective of general \textit{constrained policy search} (CPS), in which the optimal policy is described as a weighted behavior policy:
\begin{equation}
    \pi^*(\bm a|\bm s)=\pi_\theta^b(\bm a|\bm s)w(\bm a|\bm s),~s.t. \int_{\mathcal A}w(\bm a|\bm s)\dd\bm a=1,~\forall \bm s,
\end{equation}
where $\pi^b_\theta(\bm a|\bm s)$ represents the behavior policy learned by the diffusion model from the dataset, and $w(\bm s,\bm a)$ is a weight function. IDQL derives its weight function from the generalized implicit Q-learning:
\begin{equation}
    w(\bm a|\bm s)=\frac{|f'(Q_\phi(\bm s, \bm a)-V^*(\bm s))|}{|Q_\phi(\bm s,\bm a)-V^*(\bm s)|},
\end{equation}
where $f$ can be any convex function, $f'=\frac{\partial f}{\partial V(\bm s)}$, and
\begin{equation}
    V^*(\bm s)=\mathop{\arg\min}\limits_{V(\bm s)}\mathbb E_{\bm a\sim\pi_\theta^b(\bm a|\bm s)}\left[f(Q_\phi(\bm s, \bm a)-V(\bm s))\right].
\end{equation}
Therefore, the training of IDQL consists of two independent processes: training the diffusion model to clone the behavior policy and training the IQL-based weight function $w(\bm a|\bm s)$. At each inference step, IDQL samples a set of candidate actions $\{\bm a_0\}$, computes the weights $\{w(\bm s, \bm a_0)\}$, and then selects the action to be executed as a categorical from $\{w(\bm s, \bm a_0)\}$.

\textbf{DiffusionBC.}~\cite{pearce2023imitating} DiffusionBC constructs an observation-to-action diffusion model for imitating stochastic and multimodal human demonstrations. The basic version of DiffusionBC applies diffusion generation directly as a diffusion policy $\pi(\bm a_0|\bm s, \bm a_{t}, t)$ with noisy action $\bm a_{t} \in \mathbb R^{|\bm a|}$, denoising timestep $t$ and observation $\bm s$ (possibly with a history) input. To better select intra-distributional actions to mimic human behavior, DiffusionBC proposed the \textbf{Diffusion-X Sampling} trick, which encourages higher likelihood actions during sampling. For diffusion-X sampling, the sampling process first runs normal $T$ denoising timesteps, and timesteps is fixed to $t=1$, then extra denoising iterations continue to run for $M$ timesteps toward higher-likelihood regions.  

\textbf{DiffusionPolicy.}~\cite{chi2023diffusion} Similar to DiffusionBC, Diffusion Policy also uses a diffusion model to directly approximate the conditional distribution $p(\bm a|\bm s)$, but uses two key design choices: (1) \textbf{Closed-loop Action-chunking Prediction}: Diffusion Policy generates sequences of actions per prediction rather than single action to encourage temporal consistency and smoothness in long-term planning to better fit multimodal distributions. At time step $t$, the policy takes the latest $T_s$ (the observation horizon) steps of observation data $\bm s_t$ as input and predicts $H$ steps of actions, of which $T_a$ (the action prediction horizon) steps of actions are executed on the robot without re-planning. (2) \textbf{Network Architecture Options}: Diffusion Policy adopts the traditional 1D-Unet~\cite{janner2022planning} and DiT~\cite{peebles2023dit} to new CNN-based Unet and time-series diffusion transformer network architectures. CNN-based Diffusion Policy conditions the action generation process on observation $\bm s$ with Feature-wise Linear Modulation (FiLM)~\cite{perez2018film} and Transformer-based Diffusion Policy fuses state $\bm s$ and action $\bm a$ features via cross attention to jointly predict $\epsilon_\theta(o, a_k, k)$, where $k$ is sinusoidal embedding for diffusion iteration. The Diffusion Policy has demonstrated excellent performance and high stability in multiple simulation environments and real-world tasks for imitation learning and is a widely used baseline for embodied AI.

\subsection{Diffusion Data Synthesizers.}
\textbf{SynthER.} \citep{lu2024synthetic} SynthER uses the diffusion model to generate one-step transitions $(\bm s, \bm a, r, d, \bm s')$. Trained on an offline dataset, SynthER then upsamples it to a larger dataset (in D4RL, SynthER upsamples each dataset to 5M transitions), which helps other offline RL algorithms to optimize the agent policy.

\section{Limitations, Challenges, and Future Directions}\label{append:limit_chall}

\textbf{Limitations.} Although the modular structure and pipeline design of \alg greatly simplify the implementation difficulty for researchers deploying DMs, the inherent complexity of the principles and improvements of DMs still requires a considerable amount of time to deeply understand each type of module. We hope to alleviate this issue and better facilitate collaboration through comprehensive configuration files and documentation, as well as active maintenance and updates. Additionally, When dealing with certain specific issues, \alg may require tailored adjustments and optimizations. For instance, the current version of \alg does not directly support discrete or hybrid action space tasks, which may be mitigated through techniques such as action representation~\cite{li2021hyar} or using categorical diffusion models~\cite{dieleman2022continuous}.

Based on experimental analyses of \alg, we have identified several promising areas for further research as follows:

\textbf{Unleashing the potential of diffusion planners.} 
Analogous to the classification of RL algorithms, as diffusion planners can imaginatively generate interactive trajectories, they should be categorized under model-based RL (MBRL). In MBRL, there are various ways to utilize learned dynamic models, including planning to search for the optimal action \citep{hafner2019planet, hao2024magumbel}, optimizing policies using rollout trajectories \citep{Hafner2020Dreamer}, and even combining these two approaches \citep{hansen2022tdmpc, hansen2024tdmpc2}. Currently, diffusion planners are limited to the first paradigm, and due to their sensitivity to guidance and lack of safety constraints, they are prone to OOD plans \citep{dong2024diffuserlite}, falling short in performance compared to other offline MBRL algorithms. Future research can explore new paradigms for diffusion planners, attempting diverse ways to utilize generated trajectories or integrating safety constraints to enhance the fidelity of generated trajectories, thereby unleashing the full potential of diffusion planners.

\textbf{Exploring the reasons behind sampling degradation.} 
In \Cref{sec:impact_of_db_ss}, we discuss an anomaly known as \textit{sampling degradation}, where the algorithm's performance decreases as the number of sampling steps increases. This anomaly has been identified in previous works \citep{kang2024efficient, chen2023sfbc} and remains an open question. Theoretically, more sampling steps should result in a more accurate SDE/ODE solution, ultimately producing higher-fidelity samples. This naturally prompts a trade-off exploration between sampling steps and performance during implementation. However, in experiments, increasing sampling steps in certain tasks does not improve performance and can even lead to a decrease. Future research can systematically investigate this anomaly to provide optimal recommendations for selecting sampling steps.

\textbf{Understanding the impact of SDE and ODE.}
In our experiments, we observe consistent differences in SDE solvers and ODE solvers on algorithm performance, tendency to sampling degradation, and sensitivity to guidance. While there is existing research on the impact of SDE and ODE in computer vision \citep{nie2024the, lu2023dpmsolverpp}, there is still a gap in research within the decision-making domain. Future research can fill this gap and explore the implications of SDE and ODE solvers in decision-making tasks.

\textbf{Accelerating Diffusion Model Sampling.} Due to the denoising process involved in iterative sampling, DMs face the issue of slow sampling speeds when used for decision-making. This poses significant challenges in scenarios such as real-time robot control or game AI. Diffuser{\footnotesize{Lite}}~\cite{dong2024diffuserlite} is a diffusion planner method that addresses this issue by modeling the diffusion process through a plan refinement process for coarse-to-fine-grained trajectory generation and further accelerates the sampling speed using rectified flow. Further speeding up the sampling speed of various roles of DMs remains a promising research direction.

\section{Potential Social Impact}\label{append:social_impact}

\alg fills a critical gap in the current landscape by providing a unified and modularized framework that empowers researchers and practitioners to explore new frontiers. This will accelerate the development and deployment of diffusion-based decision-making applications, such as various robotics research and products. However, \alg may also be used in military weapon development.

\section{License}
\label{append:licence}
Our codebase is released under Apache License 2.0.

\newpage
\section*{Checklist}

\begin{enumerate}

\item For all authors...
\begin{enumerate}
  \item Do the main claims made in the abstract and introduction accurately reflect the paper's contributions and scope?
    \answerYes{See the abstract and \Cref{sec:introduction}.}
  \item Did you describe the limitations of your work?
    \answerYes{See \Cref{append:limit_chall}}
  \item Did you discuss any potential negative societal impacts of your work?
    \answerYes{See \Cref{append:social_impact}}
  \item Have you read the ethics review guidelines and ensured that your paper conforms to them?
    \answerYes{We read the ethics review guidelines and ensured that our paper conforms to them.}
\end{enumerate}

\item If you are including theoretical results...
\begin{enumerate}
  \item Did you state the full set of assumptions of all theoretical results?
    \answerNA{We are including no theoretical results.}
	\item Did you include complete proofs of all theoretical results?
    \answerNA{We are including no theoretical results.}
\end{enumerate}

\item If you ran experiments (e.g. for benchmarks)...
\begin{enumerate}
  \item Did you include the code, data, and instructions needed to reproduce the main experimental results (either in the supplemental material or as a URL)?
    \answerYes{We release the code and include instructions in the supplemental material and our \href{https://github.com/CleanDiffuserTeam/CleanDiffuser}{project website}.}
  \item Did you specify all the training details (e.g., data splits, hyperparameters, how they were chosen)?
    \answerYes{See \Cref{append:hyperparams}.}
	\item Did you report error bars (e.g., with respect to the random seed after running experiments multiple times)?
    \answerYes{We report the mean and standard error over 150 episode seeds in all our experiments.}
	\item Did you include the total amount of compute and the type of resources used (e.g., type of GPUs, internal cluster, or cloud provider)?
    \answerYes{See \Cref{append:computing_resources}}
\end{enumerate}

\item If you are using existing assets (e.g., code, data, models) or curating/releasing new assets...
\begin{enumerate}
  \item If your work uses existing assets, did you cite the creators?
    \answerYes{We have cited all creators and works corresponding to the assets we have used.}
  \item Did you mention the license of the assets?
    \answerYes{We have mentioned the licenses of all the benchmarks and datasets that we have used. See \Cref{append:env_details} and \Cref{append:licence}.}
  \item Did you include any new assets either in the supplemental material or as a URL?
    \answerYes{We release and open-source \alg, which includes many new assets.}
  \item Did you discuss whether and how consent was obtained from people whose data you're using/curating?
    \answerYes{See \Cref{append:env_details}}
  \item Did you discuss whether the data you are using/curating contains personally identifiable information or offensive content?
    \answerYes{See \Cref{append:env_details}}
\end{enumerate}

\item If you used crowdsourcing or conducted research with human subjects...
\begin{enumerate}
  \item Did you include the full text of instructions given to participants and screenshots, if applicable?
    \answerNA{We did not use crowdsourcing or conduct research with human subjects.}
  \item Did you describe any potential participant risks, with links to Institutional Review Board (IRB) approvals, if applicable?
    \answerNA{We did not use crowdsourcing or conduct research with human subjects.}
  \item Did you include the estimated hourly wage paid to participants and the total amount spent on participant compensation?
    \answerNA{We did not use crowdsourcing or conduct research with human subjects.}
\end{enumerate}

\end{enumerate}

\end{document}